\documentclass[letterpaper]{article} % DO NOT CHANGE THIS
\usepackage{aaai2026}  % DO NOT CHANGE THIS
\usepackage{times}  % DO NOT CHANGE THIS
\usepackage{helvet}  % DO NOT CHANGE THIS
\usepackage{courier}  % DO NOT CHANGE THIS
\usepackage[hyphens]{url}  % DO NOT CHANGE THIS
\usepackage{graphicx} % DO NOT CHANGE THIS
\usepackage{natbib}  % DO NOT CHANGE THIS AND DO NOT ADD ANY OPTIONS TO IT
\usepackage{caption} % DO NOT CHANGE THIS AND DO NOT ADD ANY OPTIONS TO IT
\usepackage{algorithm}
\usepackage{algorithmic}
\usepackage{array}
\usepackage{multirow}
\usepackage{amsmath}
\usepackage{amssymb}
\usepackage{dsfont}
\usepackage{booktabs}
\usepackage{subcaption}
\usepackage[T1]{fontenc}

\usepackage{newfloat}
\usepackage{listings}
\DeclareCaptionStyle{ruled}{labelfont=normalfont,labelsep=colon,strut=off} % DO NOT CHANGE THIS
\floatstyle{ruled}
\newfloat{listing}{tb}{lst}{}
\floatname{listing}{Listing}
\title{DiM-TS: Bridge the Gap between Selective State Space Models and Time \protect\\ Series for Generative Modeling}
\author{
    Zihao Yao\textsuperscript{\rm 1},
    Jiankai Zuo\textsuperscript{\rm 1},
    Yaying Zhang\textsuperscript{\rm 1}\thanks{Corresponding author.}
}
\affiliations{
    \textsuperscript{\rm 1}The Key Laboratory of Embedded System and Service Computing, Ministry of Education,\\

    Tongji University, Shanghai 200092, China\\
    \{yaozihao, tj\_zjk, yaying.zhang\}@tongji.edu.cn
}

\begin{document}

\maketitle

\begin{abstract}
Time series data plays a pivotal role in a wide variety of fields but faces challenges related to privacy concerns. Recently, synthesizing data via diffusion models is viewed as a promising solution. However, existing methods still struggle to capture long-range temporal dependencies and complex channel interrelations.
In this research, we aim to utilize the sequence modeling capability of a State Space Model called Mamba to extend its applicability to time series data generation. 
We firstly analyze the core limitations in State Space Model, namely the lack of consideration for correlated temporal lag and channel permutation. Building upon the insight, we propose Lag Fusion Mamba and Permutation Scanning Mamba, which enhance the model's ability to discern significant patterns during the denoising process. 
Theoretical analysis reveals that both variants exhibit a unified matrix multiplication framework with the original Mamba, offering a deeper understanding of our method.
Finally, we integrate two variants and introduce Diffusion Mamba for Time Series (DiM-TS), a high-quality time series generation model that better preserves the temporal periodicity and inter-channel correlations.
Comprehensive experiments on public datasets demonstrate the superiority of DiM-TS in generating realistic time series while preserving diverse properties of data.
\end{abstract}

\begin{links}
    \link{Code}{https://github.com/yzh8221/DiMTS}
\end{links}

\section{Introduction}

Time series data has been extensively applied in diverse domains for effective data analysis and prediction tasks, including finance, energy and climate \cite{monash}.
However, privacy concerns frequently hinder the data collection process, limiting the accessibility of real-world data \cite{privacy}.
Additionally, in data-scarce domains like energy, the requirement for rich and high-quality datasets is challenging \cite{pad-ts}.
To address above issues, synthesizing realistic time series that closely resemble but do not replicate the original dataset has emerged as a promising solution, attracting increasing attention in recent years.
Due to the superior training stability compared to GANs and higher-quality samples than VAEs, denoising diffusion probabilistic models (DDPMs) \cite{ddpm} have become the prevailing paradigm in generative modeling \cite{diffusion-ts}.

Despite advancements, the Transformer-based architectures in most existing methods remain susceptible to noise \cite{crossgnn}, which may generate unrealistic distribution during the denoising process. Furthermore, the self-attention mechanism is inherently permutation-invariant \cite{dlinear}, leading the model to perform numerical approximation rather than capturing temporal dependencies. As a result, the synthetic samples suffer from low quality, as essential temporal properties are not explicitly preserved.

Meanwhile, State Space Models (SSMs) demonstrate great potential for long sequence modeling \cite{mamba2}. Among them, Mamba \cite{mamba} has recently gained popularity due to its selection mechanism that parameterize input tokens to filter out irrelevant information. 
Despite the inherent suitability of SSMs for modeling time series, their integration into generation task remains largely underexplored. It is mainly hindered by two key challenges.

\begin{figure}
    \centering
    \includegraphics[width=\linewidth]{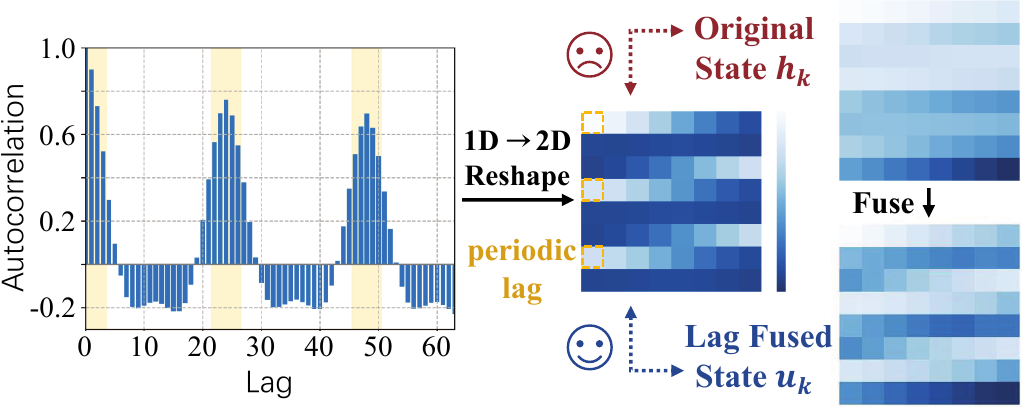}
    \caption{Comparison of ACF values, the latent state in original SSMs, and the lag fused state. We reshape them into a 2D format for clearer presentation. While the original latent state fails to capture periodic dependencies observed in ACF values, the lag fused state performs better in this regard.}
    \label{state_visual}
\end{figure}

\textbf{(1) The lack of inductive bias for modeling correlated lags in temporal dimension.}
As shown in Figure \ref{state_visual}, the autocorrelation function (ACF) typically exhibits high values at fixed periodic intervals, reflecting the similarity and dependency between the current time step and specific lag. While SSMs outperform Transformers in capturing temporal dynamics, the inherent unidirectional scanning paradigm inevitably ignore such correlations.
The latent state of SSMs in Figure \ref{state_visual} reveals unidirectional attenuation along the temporal scanning, contradicting the variation pattern of the ACF. The inconsistency disrupts the temporal semantics associated with periodicity, hindering the model's ability to preserve temporal dependencies during the denoising process.

\textbf{(2) The difficulty of capturing complex variable interactions in channel dimension.}
Channel correlation is crucial for time series, as the modeling of a particular channel can be enhanced by leveraging information from related channels.
However, global attention is susceptible to interference from irrelevant channels, hindering the recovery of lost inter-channel dependencies from noise.
While Mamba with selection mechanism offering a potential solution, it tends to shift focus toward recent input \cite{spatial-mamba}. As a result, highly correlated channels are insufficiently modeled if they are distant in scanning order. This highlights the need for effective time series channel permutation strategy.

In this study, we tackle the aforementioned challenges by presenting DiM-TS, a novel diffusion model that pioneers bridging the gap between selective SSMs and time series for generative modeling. It adopts an encoder-based dual-channel architecture to better capture time series properties across multiple dimensions.
As the core technique, we propose Lag Fusion Mamba for temporal denoising and Permutation Scanning Mamba for channel denoising.
The former fuses latent state of SSMs with correlated lags, introducing inductive bias to explicitly model periodicity while preserving the temporal dynamics as in Figure \ref{state_visual}.
The latter introduces a correlation-aware permutation strategy that leverages the attention shift of Mamba to enhance modeling between highly correlated channels.
We further show that both modules and original Mamba can be represented within a unified structured matrix framework, offering a clearer conceptual understanding of our method.
Additionally, we design a multi-feature loss to reconstruct samples rather than noises in each diffusion step, which encourage samples to approach realistic distribution from multiple perspectives.

Our contributions are summarized as follows.
\begin{itemize}
    \item We present DiM-TS that better leverages the advantages of SSMs in time series generation. To the best of our knowledge, we are the first to  bridge the gap between selective SSMs and time series for generative modeling.
    \item Motivated by the limitations of SSMs in modeling temporal dependencies and channel correlation, we propose two effective variants: Lag Fusion Mamba and Permutation Scanning Mamba. We further prove their unification with Mamba under the structured matrix framework.
    \item Experiments under challenge settings demonstrate that DiM-TS achieves superior performance in generating time series that preserve multiple key properties.
\end{itemize}

\section{Related Work}

\subsection{Generative Models in Time Series}

Generative Adversarial Networks (GANs) \cite{gan, c_rnn_gan}, which jointly optimize generator and discriminator, have been widely applied to time series generation \cite{psa-gan} but suffer from training instability. VAE-based models \cite{timeVAE, vae_bayes} enable fast and diverse sampling, yet often produce low-quality samples and struggle with KL divergence optimization \cite{freq-diff}.
Denoising diffusion probabilistic models (DDPMs) \cite{ddpm} emerge as a new class of generative framework and have demonstrated effectiveness in domains like images \cite{zigma} and trajectory \cite{difftraj}. Recently, diffusion models have also been developed for time series. Diffusion-TS \cite{diffusion-ts} improves generalization and interpretability by disentangling temporal components such as trend and seasonality. PaD-TS \cite{pad-ts} explicitly considers time series population-level property preservation overlooked by previous approaches. 
Despite the advancements, Transformer architecture adopted by most methods are inherently time-invariant \cite{dlinear}. This hinders the preservation of essential temporal properties, thereby degrading the fidelity and quality of generated samples.

\subsection{State Space Models}

SSMs are mathematical framework depicting the system dynamic behavior over time \cite{ssm_intro}. LSSL \cite{lssl} connects SSMs with Recurrent models and introduces the HiPPO \cite{hippo} framework to handle long term dependencies. To mitigate resource scarcity issue, S4 \cite{s4} leverages structured SSMs to improve the efficiency and scalability. However, the linear time invariance formulation limits the context-awareness. 
To this end, Mamba \cite{mamba} introduces a hardware-efficient selection mechanism that filters noise and propagates relevant information by parameterizing the input. Researchers further adapt Mamba to domain-specific requirements. Spatial-Mamba \cite{spatial-mamba} utilizes dilated convolutions to capture image spatial structural dependencies. ZigMa \cite{zigma} integrates a continuous scanning scheme with DDPMs for visual data generation.

Despite the effectiveness of Mamba that have demonstrated across various domains, its application for generative time series modeling remains unexplored. 
In this work, we aim to address the limitations of Mamba for time series data to fully exploit its potential and bridge this gap.

\begin{figure*}[t]
\centering
\includegraphics[width=\textwidth]{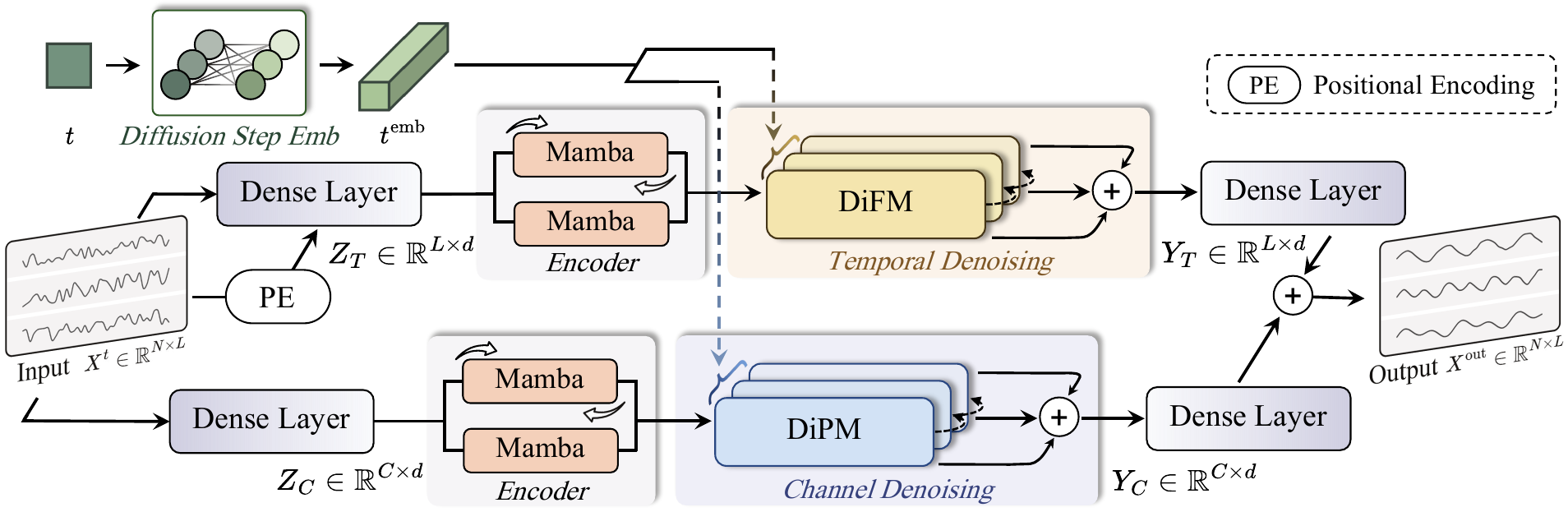}
\caption{Proposed DiM-TS framework. Diffusion State Fusion Mamba (DiFM) and Diffusion Scanning Permutation Mamba (DiPM) are tailored for temporal denoising and channel denoising during generation process, respectively.}
\label{main}
\end{figure*}

\section{Preliminaries}

We begin by presenting the definition of time series generation, then, we briefly review the formulations of DDPMs and SSMs.
Please refer to Appendix B and D for details.

\subsection{Problem Statement}

Given observations of a multivariate time series dataset $\mathcal{D}=\{x_i\}_{i=1}^{M}$ with $M$ samples. Each sample $x_i\in \mathbb{R}^{L\times C}$ is a multivariate time series, where $L$ is the sequence length and $C$ denotes the number of channels. Our unconditional generation goal is utilizing diffusion-based models to map Gaussian noise to a synthetic dataset $\mathcal{D}_{\text{syn}} = \{\bar{x}_i\}_{i=1}^{M}$ that approximates the distribution of original dataset $\mathcal{D}$.

\subsection{Denoising Diffusion Probabilistic Models}

Diffusion models are a type of generative model that contain forward process and reverse process. The forward process is a Markov process where a sample $x^0 \sim q(x)$ is gradually noised into standard Gaussian noise $x^T \sim \mathcal{N}(0, \mathbf{I})$ by incrementally adding noise at each diffusion step $t$:
\begin{equation}
    q\left(\mathbf{x}^{t} \mid \mathbf{x}^{t-1}\right)=\mathcal{N}\left(\mathbf{x}^{t} ; \sqrt{1-\beta^{t}} \mathbf{x}^{t-1}, \beta^{t} \mathbf{I}\right),
\end{equation}
where $t\in[1,T]$, $\beta^{t}\in(0,1)$. The reverse process gradually denoise samples via reverse transitions:
\begin{equation}
    p_{\theta}\left(\mathbf{x}^{t-1} \mid \mathbf{x}^{t}\right)=\mathcal{N}\left(\mathbf{x}^{t-1} ; \mu_{\theta}\left(\mathbf{x}^{t}, t\right), \Sigma_{\theta}\left(\mathbf{x}^{t}, t\right)\right),
\end{equation}
where $\mu_{\theta}(\cdot)$ is a learnable parameter, $\sum_{\theta}(\cdot)$ is fixed as $\sigma_t^2\mathbf{I}$.

The reverse process can be reduced to learning a surrogate approximator to parameterize $\mu_{\theta}(x^t,t)$ for all $t$. Hence, the denoising model parameters $\theta$ are optimized by minimizing:
\begin{equation}
\mathcal{L}_0(\theta)=\sum_{t=1}^{T}\mathbb{E}_{q(x^t|x^0)}||\mu(x^t,x^0)-\mu_{\theta}(x^t,t)||,
\end{equation}
where $\mu(x^t,x^0)$ is the mean of posterior $q(x^{t-1}|x^0,x^t)$.

\subsection{State Space Models}

SSMs are typically linear time-invariant system mapping input $x(k) \in\mathbb{R}^H$ to $y(k) \in \mathbb{R}^H$ via latent state $h(k)\in \mathbb{R}^{{N}\times{H}}$. This dynamic system can be described by the linear state transition and observation equations as:
\begin{equation}\label{LTI}
\begin{aligned}
h^{\prime}(k)=\mathbf{A}h(k)+\mathbf{B}x(k), \quad
y(k)=\mathbf{C}h(k)+\mathbf{D}x(k).
\end{aligned}
\end{equation}
$\mathbf{A} \in \mathbb{R}^{N\times N}$ is state transition matrix. $\mathbf{B}\in\mathbb{R}^{N\times 1}, \mathbf{C}\in \mathbb{R}^{1\times N}$ are projection parameters. $\mathbf{D}\in\mathbb{R}$ is typically omitted (assume $\mathbf{D}=0$), as it can be viewed as a skip connection.

Due to the hardness of analytical solutions for solving Eq. (\ref{LTI}), ZOH \cite{zoh} is applied to approximate the continuous-time SSMs into a discrete-time form. Given a parameter $\Delta$, the discretized SSMs can be represented as:
\begin{equation}\label{eq_o}
\begin{aligned}
h_k=\mathbf{\bar{A}}_kh_{k-1}+\mathbf{\bar{B}}_kx_k, \quad
y_k=\mathbf{C}_kh_k,
\end{aligned}
\end{equation}
where $\mathbf{\bar{A}}=\text{exp}(\Delta \mathbf{A})$, $\mathbf{\bar{B}}=(\Delta \mathbf{A})^{-1}(\text{exp}(\Delta \mathbf{A})-\mathbf{I})\cdot \Delta \mathbf{B}$.
Since time-independent parameters lack content-aware representation, Mamba introduces the selective mechanism that makes $\mathbf{B},\mathbf{C}$ and $\Delta$ depend on input $x_t$.

\section{Methodology}

\subsection{Model Framework}

The architecture of DiM-TS is depicted in Figure \ref{main}. It comprises two parts: temporal dependencies modeling and channel interactions modeling. Each part utilize an encoder-decoder module to capture time series patterns. The representations are subsequently fused to obtain the final output.

\subsubsection{Embedding}

Given a diffusion step $t$ and its corresponding noised time series $x^t\in\mathbb{R}^{L\times C}$, we obtain temporal first input $x_T \in \mathbb{R}^{L\times C}$ and channel first input $x_C \in \mathbb{R}^{C\times L}$ by permuting $L$ and $C$ separately. Then, $x_T$ and $x_C$ pass through linear dense layer to learn context representations:
\begin{equation}
    z_T=(W_T^1x_T+b_T^1)+\text{PE},\quad z_C=W_C^1x_C+b_C^1
\end{equation}
where $W_T^1,W_C^1,b_T^1,b_C^1$ are learnable parameters. \text{PE} denotes an additional positional encoding, where $\text{PE}_{pos,2i}=\sin(\frac{pos}{10000^{2i/d}})$, $\text{PE}_{pos,2i+1}=\cos(\frac{pos}{10000^{2i/d}})$.

\subsubsection{Encoder}

Selection mechanism has demonstrated the effectiveness in filtering out irrelevant information. However, the unidirectional scanning paradigm can only incorporate preceding input.
Here, we utilize two vanilla Mamba to form a bidirectional Mamba encoder layer $\text{Bi-Mamba}(\cdot)$ to extract relative features at each diffusion step:
\begin{equation}
    Z_T=\text{Bi-Mamba}(z_T), \quad Z_C=\text{Bi-Mamba}(z_C).
\end{equation}

\subsubsection{Decoder}
Since DiT \cite{DiT} has been validated as an effective diffusion framework in high throughput and condition incorporating, we mirror the backbone of DiT to devise Diffusion State Fusion Mamba (DiFM) and Diffusion Scanning Permutation Mamba (DiPM) as the final layers in DiM-TS (see Appendix C for details).
The diffusion timestep $t$, serving as conditional information, is transformed to $t^{\text{emb}}$ via dense layers and then incorporated into each denoising layer. Given the encoded $Z_T$ from previous section, the generation process can be formally described as:
\begin{equation}
\begin{aligned}
    Y^{i}_T=& \mathds{1}_{i=0} \text{DiFM}(Z_T, t^{\text{emb}})+\mathds{1}_{i=1} \text{DiFM}(Y^{0}_T+Z_T, t^{\text{emb}})\\
    & +\mathds{1}_{i>0} \text{DiFM}(Y^{i-1}_T+Y^{i-2}_T, t^{\text{emb}}),
\end{aligned}
\end{equation}
where $Y^{i}_T$ is the output of $i^{\text{th}}$ \text{DiFM} block. $\mathds{1}_{c}$ represents the indicator function, which evaluates to $1$ if the condition $c$ holds, and $0$ otherwise. Subsequently, the temporal representation $Y_{T}$ can be learned by adding the output of all \text{DiFM} blocks.
For the channel dimension, we simply replace $Z_T$ with $Z_C$ and apply the same procedure to obtain $Y_C$.

Eventually, we convert the temporal and channel representation to their original shape with dense layers, and obtain the final output through summation:
\begin{equation}
    x^{\text{out}}(x^t,t,\theta)=(W_T^2Y_T+b_T^2)+(W_C^2Y_C+b_C^2),
\end{equation}
where $W_T^2,W_C^2,b_T^2,b_C^2$ are learnable parameters.

\subsection{Training Objective}

In the reverse denoising process, the model is trained to generate time series via the following objective function:
\begin{equation}
    \mathcal{L}_{\text{DDPM}}=\mathbb{E}_{t,x_0}[\|x^0-x^{\text{out}}(x^t,t,\theta)\|^2].
\end{equation}
However, the training loss solely focuses on the authenticity of data at the individual level, neglecting higher-level statistical properties \cite{pad-ts}. For instance, traffic flow typically exhibits periodic peak patterns, and weather data often contains correlated fluctuations between pressure and humidity.
To address this, we introduce additional multi-feature loss to guide the diffusion process.

Since most temporal information is localized on the low frequencies, imposing constraint in the frequency domain can enhance sample fidelity by preserving underlying temporal property \cite{fourier-diff}.
Fourier transform that converts time domain signal to frequency domain representation has proven to be an effective operation \cite{diffusion-ts}. The Fourier-based auxiliary loss can be defined as:
\begin{equation}
    \mathcal{L}_{T}=\|\mathcal{FFT}(x^0)-\mathcal{FFT}(x^{\text{out}}(x^t,t,\theta))\|^2,
\end{equation}
where $\mathcal{FFT}(\cdot)$ denotes the Fast Fourier Transformation.

Meanwhile, to capture the correlation distribution shift in channel dimension, we adopt Maximum Mean Discrepancy (MMD) inspired by \cite{pad-ts}. By calculating all $P=\frac{C(C-1)}{2}$ pairwise channel correlation distribution shift, the regularization loss can be defined as:
\begin{equation}
    \mathcal{L}_C=\frac{1}{P}\sum_{i=1}^{P}\text{MMD}(D^{i}(x^0),D^{i}(x^{\text{out}})),
\end{equation}
$D^i$ denotes the correlation distribution of $i^{\text{th}}$ pair channels.
We further employ the Same Diffusion Step Sampling strategy \cite{pad-ts} for reasonable distribution comparison.

Hence, the training objective can be formulated as:
\begin{equation}
\mathcal{L}=\mathcal{L}_{\text{DDPM}}+\lambda_1\mathcal{L}_{T}+\lambda_2\mathcal{L}_{C},
\end{equation}
where $\lambda_1$ and $\lambda_2$ are hyperparameter to balance loss terms.

\subsection{Enhanced Mamba Module}

\begin{figure}[t]
\centering
\includegraphics[width=\linewidth]{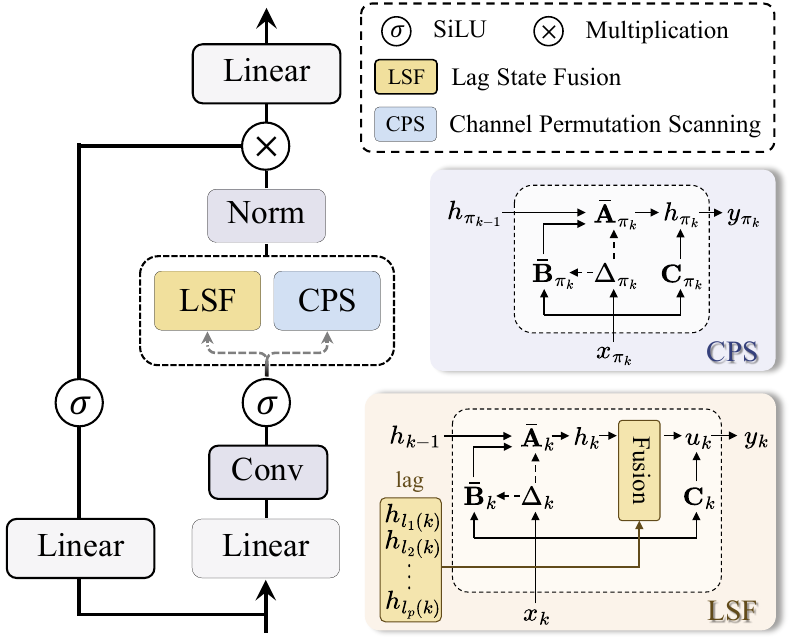}
\caption{Architecture of proposed Mamba variants. Lag Fusion Mamba employs the LSF equation, while Permutation Scanning Mamba adopts the CPS equation.}
\label{ssm}
\end{figure}

In this section, we present the proposed Lag Fusion Mamba and Permutation Scanning Mamba. They are incorporated as core components into DiFM and DiPM, respectively.

\subsubsection{Lag Fusion Mamba}
It is designed to capture the temporal dependencies of lag inspired by the characteristic of autocorrelation. Here, we introduce Lag State Fusion (LSF) equation into SSMs formula as shown in Figure \ref{ssm}.
It fuses lag features in latent state space without disrupting the inherent sequential nature of time series and facilitates cross-temporal modeling.
The process is formulated as:
\begin{equation}\label{ssm_f}
\begin{aligned}
h_k&=\mathbf{\bar{A}}_kh_{k-1}+\mathbf{\bar{B}}_kx_k,\\
u_k&=\sum_{p\in \Omega}\eta_p h_{l_p(k)},\quad y_k=\mathbf{C}_ku_k,
\end{aligned}
\end{equation}
where $h_k$ is the original latent state, $u_k$ is the lag fusion state, $\Omega$ is the lag set, $\eta_p$ is a learnable weight, and $l_p(k)$ is the index of the $p^{\text{th}}$ lag of position $k$.
Compared with the original Mamba where the state $h_{k}$ is solely influenced by previous state $h_{k-1}$, the state $u_{k}$ is combined with additional lag state through linear weighting fusion, resulting in a richer representation of both local and long-term temporal semantics.
Moreover, an inductive bias of correlated lags is introduced to latent state prior to projecting, which enables DiFM capable of capturing periodic dependencies from noised data. Specifically, when the lag set $\Omega$ contains only the current state $h_{k}$, the formulation reduces to the original Mamba.

In practice, considering the regular time intervals and fixed lag, we implement linear weighted state fusion via multi-scale dilated convolutions.
The 1D state sequence is first reshaped into 2D and processed by depth-wise convolutions with varying dilation factors defined by $\Omega$.
Finally, the fused 2D state is flattened to generate the output.

\subsubsection{Permutation Scanning Mamba}
It aims to scan time series channels using a coherent permutation. To this end, we augment SSMs by introducing Channel Permutation Scanning (CPS) equation as depicted in Figure \ref{ssm}. It rearranges tokens before feeding into Eq. (\ref{eq_o}). For a given permutation $\pi=\{\pi_1,\pi_2,\ldots,\pi_C\}$, the process can be expressed as:
\begin{equation}\label{ssm_p}
\begin{aligned}
h_{\pi_k}=\mathbf{\bar{A}}_{\pi_k}h_{\pi_{k-1}}+\mathbf{\bar{B}}_{\pi_k}x_{\pi_k},\quad
y_{\pi_k}=\mathbf{C}_{\pi_k}h_{\pi_k},
\end{aligned}
\end{equation}
where $\pi_k$ denotes the index of the $k^{\text{th}}$ scanning token. 

Since inter-channel correlations reflect consistent variation patterns, preserving the proximity of highly correlated channels during scanning facilitates more accurate noise estimation and synthetic data with realistic inter-channel dependencies.
Based on the observation, we propose a permutation strategy that keep related channels adjacent while separating unrelated channels apart during scanning.
We take channel similarity matrix $\mathbf{G}$ derived from arbitrary metric (e.g., Pearson) as input. Each channel can be represented by $v \in \mathbb{R}$ for sequence order, $g_{ij} \in \mathbf{G}$ represents the closeness between channel $i$ and $j$. To approximate the difference between $v_i$ and $v_j$ with $g_{ij}$, we optimize following function:
\begin{equation}\label{laplace}
\begin{aligned}
\min \sum_{i=1}^{C}\sum_{j=1}^{C}\left\|v_{i}-v_{j}\right\|^{2} g_{ij}.
\end{aligned}
\end{equation}
After obtaining vector $V=(v_1,v_2,\ldots,v_C)$, the ordered vector $V_{\pi}=(v_{\pi_1},v_{\pi_2},\ldots,v_{\pi_C})$ can be generated by sorting element values. Since the value of correlated channels are numerically close, their permutation in $V_{\pi}$ is also adjacent, thus yielding the desired permutation $\pi=\{\pi_1,\ldots,\pi_C\}$.

Channel order rearrangement can be implemented by transformation matrix $H$ defined on $\pi$, where $H_{ij}=\mathds{1}_{i=\pi_j}$.
Specially, when $\pi=\{1,2,\ldots,C\}$, $H$ reduces to the identity, and CPS coincides with the original SSMs.

In practice, the input $x$ is first transformed into $Hx$. Following processing with Eq. (\ref{ssm_p}), the output is standardized to the original permutation by inverse transformation $H^{-1}$.

\subsection{Connection with Original Mamba}

In this section, we conduct an in-depth analysis of the interrelations among original Mamba and proposed variants, aiming to elucidate the underlying mechanisms of our approach
(see Appendix D for more detailed derivations).

\subsubsection{Mamba}
The formulation is defined in Eq. (\ref{eq_o}). We can derive the latent state $h_t$ by induction:
\begin{equation}
h_k=\bar{\mathbf{A}}_k\ldots\bar{\mathbf{A}}_1h_0+\ldots+\bar{\mathbf{A}}_k\bar{\mathbf{B}}_{k-1}x_{k-1}+\bar{\mathbf{B}}_kx_k.
\end{equation}
By multiplying with $\mathbf{C}_k$ to produce $y_k$, vectorizing the equation over $k$, and setting the initial latent state $h_0=\bar{\mathbf{{B}}}_0x_0$, we establish the matrix transformation form of SSMs:
\begin{equation}\label{unifty_o}
y_k=\sum_{i=0}^{k}\mathbf{C}_k^{\top}\bar{\mathbf{A}}_{i:k}^{\times}\bar{\mathbf{B}}_{i}x_i, \quad y=\text{SSM}(x)=\mathbf{M}x,
\end{equation}
where $\mathbf{M}$ is a lower triangular matrix, $\mathbf{M}_{ki}:=\mathbf{C}_k^{\top}\bar{\mathbf{A}}_{i:k}^{\times}\bar{\mathbf{B}}_{i}$, $\bar{\mathbf{A}}_{i:k}^{\times}:=\Pi_{j=i+1}^{k}\bar{\mathbf{A}}_j$ denotes the product of the state transition matrices indexed from $i + 1$ to $k$ for $i < k$, and is defined as identity matrix when $i=k$.

\subsubsection{Lag Fusion Mamba}
Under the same setting as Eq. (\ref{ssm_f}) and Eq. (\ref{unifty_o}), the fused latent state and corresponding output in LSF equation can be reformulated as follows:
\begin{equation}
    u_k=\sum_{p\in\Omega}\sum_{i\leq l_p(k)}\eta_p\bar{\mathbf{A}}_{i:l_p(k)}^{\times}\bar{\mathbf{B}}_{i}x_i,\text{  } y=\text{LSF}(x)=\mathbf{M}^{F}x,
\end{equation}
$\mathbf{M}^F$ is adjacency matrix, $\mathbf{M}^F_{ki}:=\sum_{p\in\Omega}\eta_p\mathbf{C}_k^{\top}\bar{\mathbf{A}}_{i:l_p(k)}^{\times}\bar{\mathbf{B}}_{i}$.

\subsubsection{Permutation Scanning Mamba}
Given the specific $H$, CPS can be rewritten based on Eq. (\ref{ssm_p}) and Eq. (\ref{unifty_o}):
\begin{equation}
    y=\text{CPS}(x)=H^{-1}\text{SSM}(Hx)=\textbf{M}^{C}x,
\end{equation}
where $\mathbf{M}^C$ can be transformed into a lower triangular matrix through row interchange, $\mathbf{M}^C:=H^{-1}\mathbf{M}H$.

\subsubsection{Analysis}
\begin{figure}[t]
        \centering
        \begin{subfigure}{0.49\linewidth}
		\centering
		\includegraphics[width=\linewidth]{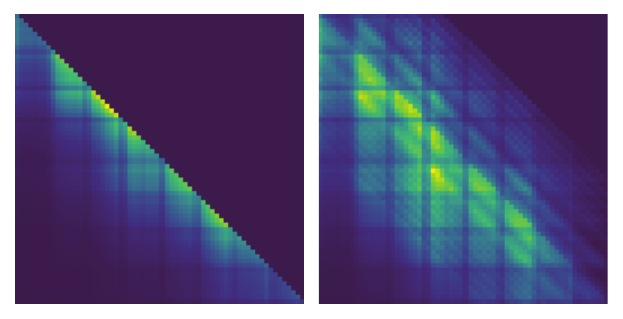}
		\caption{$\mathbf{M}$ (left) and $\mathbf{M}^F$ (right).}
		\label{MT}
	\end{subfigure}
	\begin{subfigure}{0.49\linewidth}
		\centering
		\includegraphics[width=\linewidth]{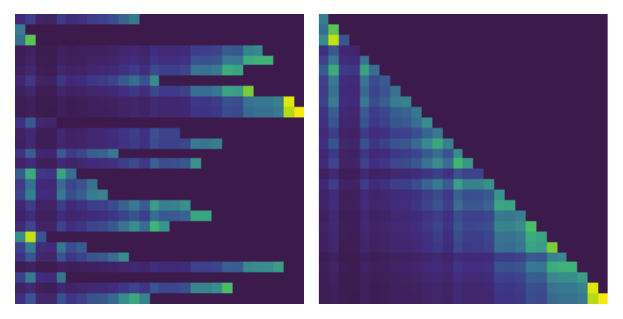}
		\caption{$\mathbf{M}^{C}$ (left) and $\mathbf{M}^{C}H$ (right).}
		\label{MC}
	\end{subfigure}
\caption{Matrix Visualizations. $\mathbf{M}$, $\mathbf{M}^{F}$, $\mathbf{M}^{C}$ denotes the matrices in Mamba, Lag Fusion Mamba and Permutation Scanning Mamba, respectively. $H$ is transformation matrix.}
\label{matrix_M}
\end{figure}

Based on the above derivation, we can conclude that all the paradigms — Mamba, Lag Fusion Mamba, and Permutation Scanning Mamba — can be modeled within a unified matrix multiplication framework, $i.e.,$ $y=\mathbf{M}x$. Their distinctions arise from the structural differences of the matrix $\mathbf{M}$.
As illustrated in Figure \ref{matrix_M}, the matrix $\mathbf{M}$ in Mamba exhibits a decaying pattern over scanning. This effect arises from the cumulative multiplication of state transition matrix $\bar{\mathbf{A}}_k$, which leads to exponential decay in attention weights as the intervals between tokens increases.
For Lag Fusion Mamba that fuses additional lag state set $\Omega$ via weighted summation, the resulting matrix $\mathbf{M}^F$ not only extents unidirectional time series modeling into a global scope, but also effectively capture high-correlated lag interactions, even when they are far apart.
Permutation Scanning Mamba, on the other hand, leverages the attention transition of Mamba while employing the channel permutation strategy. This makes model shift focus toward previously high-correlated channels and suppress attention to irrelevant ones.

\section{Experiments}

In this section, we describe the experiment settings and evaluate DiM-TS across various domains and sequence lengths. We also provide visualization results to enhance the understanding of our model behavior. Finally, we conduct an ablation study to assess the effectiveness of components.

\begin{table*}[t]
\centering
\begin{tabular}{c|c|c|c|c|c}
\toprule
Metric                                                                                               & Methods          & Stocks                     & ETTh                                        & Energy                                      & KDD-Cup                    \\ \midrule
\multirow{6}{*}{\begin{tabular}[c]{@{}c@{}}Context-FID\\ score\\ (Lower the Better)\end{tabular}}    & DiM-TS           & \textbf{0.0440$\pm$0.0074} & \textbf{0.0259$\pm$0.0021}                  & \textbf{0.0320$\pm$0.0003}                  & \textbf{0.0220$\pm$0.0029} \\
                                                                                                     & PaD-TS           & {\underline {0.0715$\pm$0.0255}}    & 1.2674$\pm$0.1881                           & {\underline {0.0657$\pm$0.0078}}                     & {\underline {0.1372$\pm$0.0208}}    \\
                                                                                                     & Diffusion-TS     & 0.4055$\pm$0.0557          & 0.2570$\pm$0.0112                           & 0.0708$\pm$0.0135                           & 1.0141$\pm$0.1186          \\
                                                                                                     & FourierDiff & 0.1294$\pm$0.0314          & {\underline {0.1198$\pm$0.0076}}                     & 0.4477$\pm$0.0467                          & 0.8294$\pm$0.1501          \\
                                                                                                     & TimeVAE          & 0.3892$\pm$0.1174          & 0.8995$\pm$0.1147                           & 3.3228$\pm$0.2680                          & 1.8987$\pm$0.2582          \\
                                                                                                     & TimeGAN          & 0.4182$\pm$0.1147          & 1.9650$\pm$0.3051                           & 1.5532$\pm$0.1681                           & 1.1560$\pm$0.3504          \\ \midrule
\multirow{6}{*}{\begin{tabular}[c]{@{}c@{}}Correlational\\ score\\ (Lower the Better)\end{tabular}}  & DiM-TS           & \textbf{0.0048$\pm$0.0029} & \textbf{0.0219$\pm$0.0046}                  & \textbf{0.4108$\pm$0.1369}                  & \textbf{3.6615$\pm$1.2297} \\
                                                                                                     & PaD-TS           & 0.0085$\pm$0.0080          & 0.1237$\pm$0.0017                           & {\underline {0.5724$\pm$0.0827}}                     & {\underline {6.7944$\pm$1.1817}}    \\
                                                                                                     & Diffusion-TS     & 0.0244$\pm$0.0053          & 0.0595$\pm$0.0053                           & 0.6360$\pm$0.0877                           & 9.4173$\pm$0.7498          \\
                                                                                                     & FourierDiff & 0.0139$\pm$0.0079          & {\underline {0.0473$\pm$0.0090}}                     & 1.1992$\pm$0.2587                           & 15.6568$\pm$2.0939         \\
                                                                                                     & TimeVAE          & 0.0859$\pm$0.0048          & 0.0593$\pm$0.0192                          & 2.1681$\pm$0.1034                           & 30.7528$\pm$1.5355         \\
                                                                                                     & TimeGAN          & {\underline {0.0059$\pm$0.0033}}    & 0.2175$\pm$0.0084                           & 3.5817$\pm$0.1221                           & 16.6840$\pm$1.5923         \\ \midrule
\multirow{6}{*}{\begin{tabular}[c]{@{}c@{}}Discriminative\\ Score\\ (Lower the Better)\end{tabular}} & DiM-TS           & \textbf{0.0291$\pm$0.0151} & \textbf{0.0053$\pm$0.0019}                  & 0.2410$\pm$0.0201                           & \textbf{0.0844$\pm$0.0233} \\
                                                                                                     & PaD-TS           & {\underline {0.0485$\pm$0.0792}}    & 0.1576$\pm$0.0137                           & \textbf{0.0919$\pm$0.0193}                  & 0.3769$\pm$0.0460          \\
                                                                                                     & Diffusion-TS     & 0.0910$\pm$0.0237          & 0.0832$\pm$0.0067                           & {\underline {0.1072$\pm$0.0162}}                     & {\underline {0.2957$\pm$0.0168}}    \\
                                                                                                     & FourierDiff & 0.0553$\pm$0.0587          & {\underline {0.0446$\pm$0.0074}}                     & 0.2062$\pm$0.0339                           & 0.4833$\pm$0.0040          \\
                                                                                                     & TimeVAE          & 0.1794$\pm$0.0801          & 0.1739$\pm$0.0935                           & 0.4999$\pm$0.0001                           & 0.4639$\pm$0.0100          \\
                                                                                                     & TimeGAN          & 0.2013$\pm$0.0712          & 0.3228$\pm$0.0738                           & 0.4995$\pm$0.0004                           & 0.4988$\pm$0.0008          \\ \midrule
\multirow{6}{*}{\begin{tabular}[c]{@{}c@{}}Predictive\\ score\\ (Lower the Better)\end{tabular}}     & DiM-TS           & \textbf{0.0367$\pm$0.0001} & $\textbf{0.1086$\pm$0.0101}$ & $\textbf{0.2474$\pm$0.0004}$ & \textbf{0.0241$\pm$0.0003} \\
                                                                                                     & PaD-TS           & {\underline {0.0368$\pm$0.0001}}    & 0.1180$\pm$0.0018                           & 0.2514$\pm$0.0002                           & {\underline {0.0282$\pm$0.0001}}    \\
                                                                                                     & Diffusion-TS     & {\underline {0.0368$\pm$0.0001}}    & 0.1173$\pm$0.0057                           & {\underline {0.2490$\pm$0.0005}}                     & 0.0324$\pm$0.0017          \\
                                                                                                     & FourierDiff & \textbf{0.0367$\pm$0.0001} & {\underline {0.1171$\pm$0.0070}}                     & 0.2508$\pm$0.0001                           & 0.0285$\pm$0.0005          \\
                                                                                                     & TimeVAE          & 0.0385$\pm$0.0003          & 0.1200$\pm$0.0044                           & 0.2888$\pm$0.0008                           & 0.0290$\pm$0.0003          \\
                                                                                                     & TimeGAN          & 0.0505$\pm$0.0007          & 0.1450$\pm$0.0046                           & 0.3129$\pm$0.0021                           & 0.0368$\pm$0.0004          \\ \bottomrule
\end{tabular}
\caption{Generation results with length 64 on multiple datasets. The best scores are in bold and the second best are underlined.}
\label{main-result}
\end{table*}

\subsection{Experiment Settings}

We briefly discuss the datasets, baselines, and evaluation metrics. All experiments are conducted on a machine with NVIDIA V100 GPU and 32GB memory.
Implementation details and code are provided in Supplementary Material.

\subsubsection{Datasets}

We utilize four major public datasets spanning diverse domains, including finance, electricity, energy and environment. (1) Stocks: Daily stock data from Google (2004-2019) with six features such as Open, Volume, etc. (2) ETTh: Electricity transformer data collected hourly, including oil temperature and six power-related metrics. (3) Energy: A UCI appliances energy prediction dataset with 28 features related to household energy consumption. (4) KDD-Cup: Hourly air quality from 2017 to 2018 estimated by 24 stations in London. More details and additional Traffic dataset results are available in Appendix F.

\subsubsection{Baselines}

We carefully select five competitive models that cover generative frameworks: (1) Diffusion-based models: PaD-TS \cite{pad-ts}, Diffusion-TS \cite{diffusion-ts}, FourierDiff \cite{fourier-diff}. (2) VAE-based model: TimeVAE \cite{timeVAE}. (3) GAN-based model: TimeGAN \cite{timeGAN}.

\subsubsection{Metrics}

The quantitative evaluation of the synthesized data is conducted from three key aspects:
1) the distribution diversity of time series. 
2) the fidelity of temporal and channel dependencies.
3) the usefulness in downstream application. 
We employ the following evaluation metrics \cite{diffusion-ts}:
(1) Context-Fréchet Inception Distance score (Context-FID score): Computes the difference between representations of real and generated data fitting into local context.
(2) Correlational score: Assess temporal dependency by absolute error between cross correlation matrices by real and generated time series.
(3) Discriminative score: Evaluates the similarity between original and generated data based on distinguishability assessed via a supervised classification model.
(4) Predictive score: Measures the usefulness of generated data by capturing Mean Absolute Error of a time series forecasting model trained on generated data.
We additionally include feature-based metrics summarized in \cite{tsgbench}: Marginal Distribution Difference (MDD), AutoCorrelation Difference (ACD), Skewness Difference (SD), Kurtosis Difference (KD). We also adopt population-level metrics from \cite{pad-ts}: Value distribution shift (VDS) and Functional dependency distribution shift (FDDS). For calculation formulas, please refer to Appendix E.

\subsection{Baselines Comparison}

\begin{table*}[t]\centering
\begin{tabular}{c|c|c|c|c|c|c}
\toprule
Metrics                                                                         & Length & DiM-TS                     & PaD-TS            & Diffusion-TS      & FourierDiff  & TimeVAE           \\ \midrule
\multirow{2}{*}{\begin{tabular}[c]{@{}c@{}}Context-FID\\ score\end{tabular}}    & 128    & \textbf{0.0451$\pm$0.0025} & 1.4856$\pm$0.2231 & 0.7100$\pm$0.0624 & 0.4753$\pm$0.0262 & 0.7571$\pm$0.0895 \\
                                                                                & 256    & \textbf{0.0516$\pm$0.0028} & 1.8520$\pm$0.2971 & 1.7604$\pm$0.0848 & 1.0262$\pm$0.0838 & 1.3814$\pm$0.1354 \\ \midrule
\multirow{2}{*}{\begin{tabular}[c]{@{}c@{}}Correlational\\ score\end{tabular}}  & 128    & \textbf{0.0215$\pm$0.0058} & 0.1171$\pm$0.0117 & 0.0890$\pm$0.0046 & 0.0855$\pm$0.0126 & 0.0556$\pm$0.0098 \\
                                                                                & 256    & \textbf{0.0225$\pm$0.0070} & 0.1357$\pm$0.0038 & 0.1144$\pm$0.0129 & 0.0916$\pm$0.0050 & 0.0444$\pm$0.0104 \\ \midrule
\multirow{2}{*}{\begin{tabular}[c]{@{}c@{}}Discriminative\\ score\end{tabular}} & 128    & \textbf{0.0033$\pm$0.0017} & 0.1707$\pm$0.0375 & 0.1436$\pm$0.0099 & 0.1619$\pm$0.0126 & 0.1835$\pm$0.0982 \\
                                                                                & 256    & \textbf{0.0044$\pm$0.0047} & 0.1979$\pm$0.0429 & 0.2103$\pm$0.0131 & 0.1939$\pm$0.1274 & 0.1984$\pm$0.0909 \\ \midrule
\multirow{2}{*}{\begin{tabular}[c]{@{}c@{}}Predictive\\ score\end{tabular}}     & 128    & \textbf{0.1115$\pm$0.0088} & 0.1270$\pm$0.0054 & 0.1122$\pm$0.0027 & 0.1175$\pm$0.0054 & 0.1151$\pm$0.0121 \\
                                                                                & 256    & \textbf{0.1050$\pm$0.0099} & 0.1106$\pm$0.0088 & 0.1171$\pm$0.0054 & 0.1150$\pm$0.0037 & 0.1163$\pm$0.0059 \\ \bottomrule
\end{tabular}
\caption{Results of long-term time series generation on ETTh dataset. The best scores are in bold.}
\label{longer_task}
\end{table*}

\subsubsection{Main Results}

\begin{figure}[t]
  \centering
  \includegraphics[width=0.98\linewidth]{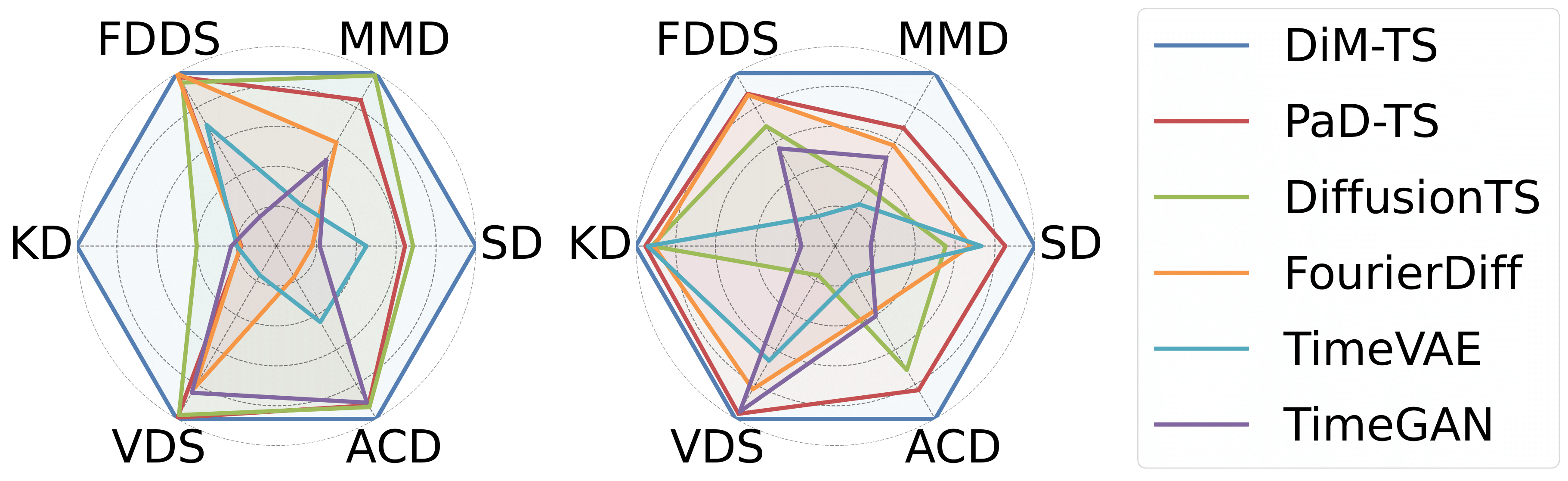}
  \caption{Feature-based and population-level measures comparison on Energy (left) and KDD-Cup (right).}
  \label{radar}
\end{figure}

We list the results of $64$-length time series generation in Table \ref{main-result}. Among the baselines concerned, DiM-TS achieves the best performance on most datasets across various metrics. It demonstrates the superiority of our method in generating high-quality synthetic time series. Notably, DiM-TS improves the context-FID score over $60\%$ and the correlation score over $35\%$ compared to previous state-of-the-art models. Moreover, the predictive score indicates that the synthetic data generated by DiM-TS is more applicable to real-world task. As shown in Figure \ref{radar}, DiM-TS achieves overall superior performance under feature-based metrics and population-level property preservation settings. The observations substantiate the capability of DiM-TS in synthesizing high-fidelity time series.

\subsubsection{Visualization}

\begin{figure}[t]
\centering
\includegraphics[width=\linewidth]{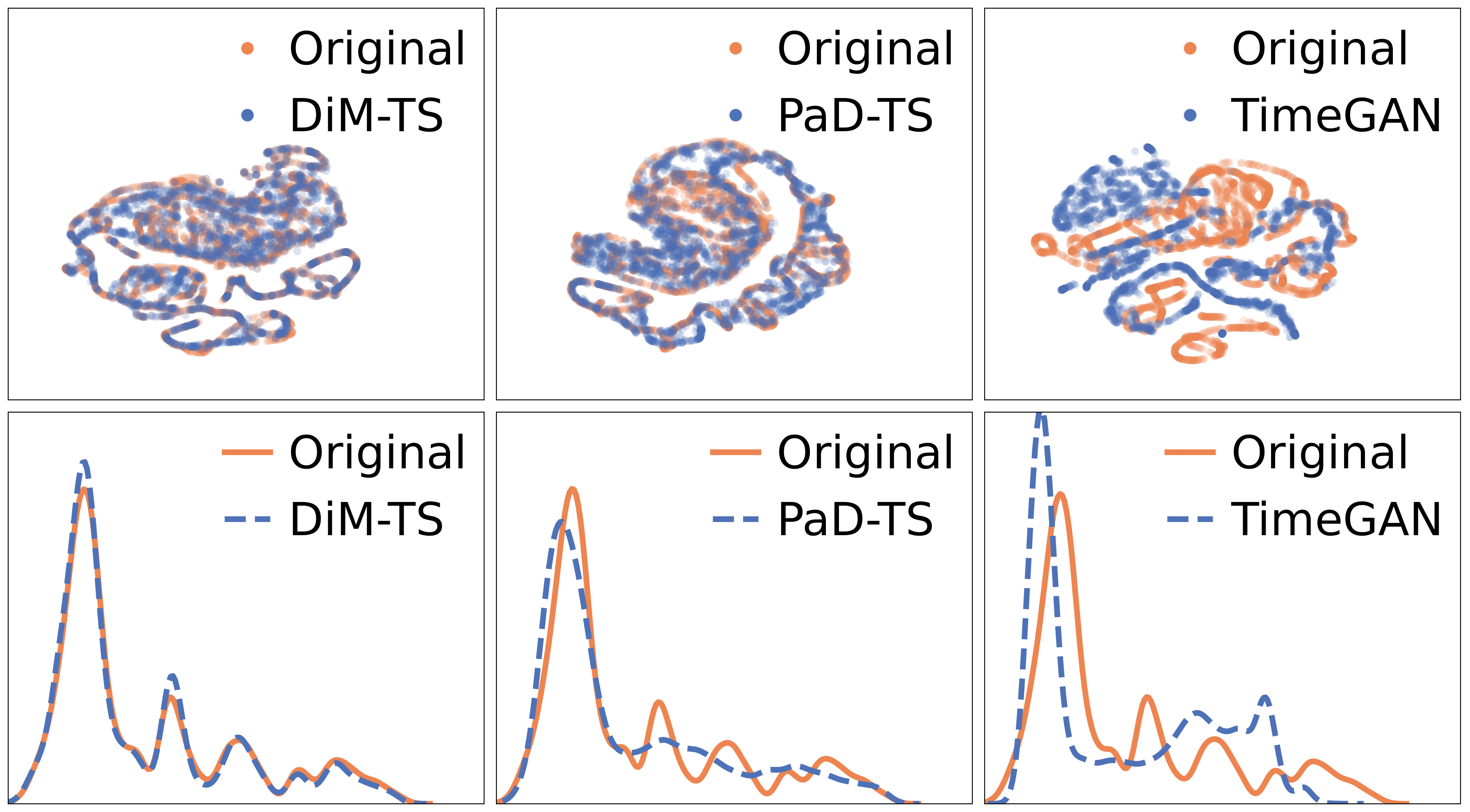}
\caption{Visualizations of t-SNE plots and data distributions on original (orange) and synthetic (blue) time series.}
\label{data_visual}
\end{figure}

To provide an intuitive understanding of model behavior, we employ the t-SNE \cite{tsne} and kernel density estimation \cite{kde} to visualize the fidelity of generated data. As shown in the $1^{st}$ row in Figure \ref{data_visual}, the 2D projection of DiM-TS using t-SNE exhibits diversity and closer alignment with the original data, whereas other methods either fail to achieve comprehensive coverage or produce unrealistic samples. The $2^{nd}$ row in Figure \ref{data_visual} shows that the synthetic time series value distribution of DiM-TS is the most similar to the original data. The results indicate that DiM-TS effectively learns the underlying statistical properties.

\subsubsection{Long sequence generation}

We further verify model stability in longer sequence generation task with lengths of $128$ and $256$. As shown in Table \ref{longer_task}, DiM-TS achieves the best performance under the challenging setting, implying the efficacy of improved components tailored for time series. Notably, the performance of DiM-TS changes steadily as the sequence length increases, demonstrating superior robustness which is meaningful for real-world applications.

\subsection{Ablation Study}

\begin{figure}[t]
\centering
\includegraphics[width=\linewidth]{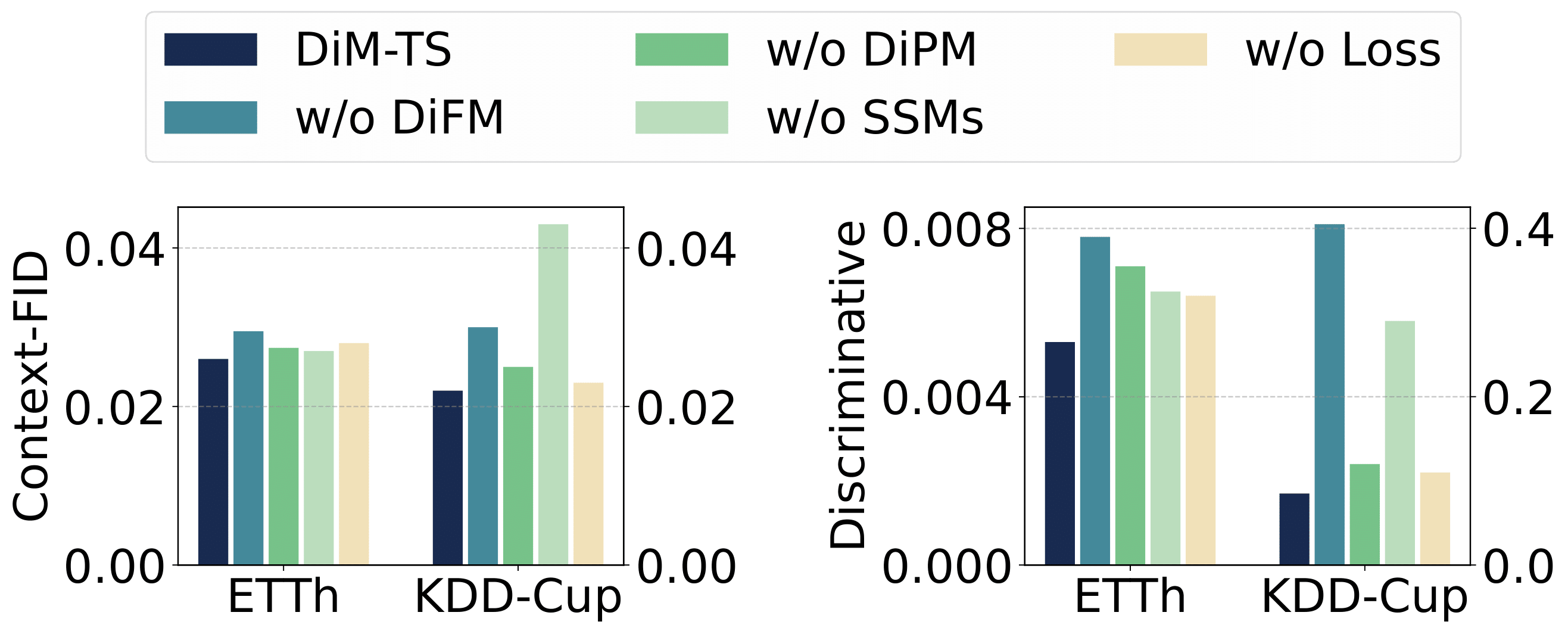}
\caption{The results of ablation study.}
\label{ab}
\end{figure}

To validate the effectiveness of components, we set up variant models for ablation experiments:
(1) w/o DiFM: The lag state fusion is removed in DiFM, i.e., it is replaced by original Mamba.
(2) w/o DiPM: The channel permutation strategy is removed in DiPM.
(3) w/o SSMs: This variant uses DiT to replace DiFM and DiPM.
(4) w/o Loss: The multi-feature loss terms $\mathcal{L}_T$ and $\mathcal{L}_C$ are omitted during training.

As shown in Figure \ref{ab}, the results highlight that Lag Fusion Mamba is the most critical part in our method, demonstrating the effectiveness of incorporating the inductive bias of correlated lags in enhancing the temporal dependency awareness of SSMs. While the Diffusion Permutation Mamba and auxiliary loss terms contribute less significantly, they still play crucial roles. Overall, integrating all components, the full DiM-TS achieves the best performance.

\section{Conclusion}

In this paper, we present DiM-TS, a novel framework that empower selective state space models for time series generation. As key contributions, we propose a Lag Fusion Mamba designed for modeling temporal dependencies, and a Permutation Scanning Mamba tailored to capturing channel correlation during the denoising process.
We further provide an in-depth analysis between the proposed variants and original Mamba, demonstrating their unification under the matrix multiplication framework and offering deeper insights into our approach.
Extensive experiments show that DiM-TS excels at synthesizing high-quality time series while preserving multiple properties across various settings.

\section{Acknowledgments}
This work was partly supported by the National Key Research and Development Program of China under Grant 2022YFB4501704, the National Natural Science Foundation of China under Grant 72342026, and Fundamental Research Funds for the Central Universities under Grant 2024-6-ZD-02.

\bibliography{aaai2026}

\begin{thebibliography}{31}
\providecommand{\natexlab}[1]{#1}

\bibitem[{Alaa and Chan(2021)}]{privacy}
Alaa, A.; and Chan, A.~J. 2021.
\newblock Generative time-series modeling with fourier flows.
\newblock In \emph{International Conference on Learning Representations}.

\bibitem[{Ang et~al.(2023)Ang, Huang, Bao, Tung, and Huang}]{tsgbench}
Ang, Y.; Huang, Q.; Bao, Y.; Tung, A.~K.; and Huang, Z. 2023.
\newblock Tsgbench: Time series generation benchmark.
\newblock \emph{arXiv preprint arXiv:2309.03755}.

\bibitem[{Candanedo(2017)}]{energy}
Candanedo, L.~M. 2017.
\newblock Data driven prediction models of energy use of appliances in a low-energy house.
\newblock \emph{Energy and buildings}, 140: 81--97.

\bibitem[{Comanescu(2012)}]{zoh}
Comanescu, M. 2012.
\newblock Integration of observer equations used in AC motor drives by zero and First Order Hold discretization.
\newblock In \emph{IECON 2012-38th Annual Conference on IEEE Industrial Electronics Society}, 3694--3698. IEEE.

\bibitem[{Crabb{\'e} et~al.(2024)Crabb{\'e}, Huynh, Stanczuk, and Van Der~Schaar}]{fourier-diff}
Crabb{\'e}, J.; Huynh, N.; Stanczuk, J.; and Van Der~Schaar, M. 2024.
\newblock Time series diffusion in the frequency domain.
\newblock \emph{arXiv preprint arXiv:2402.05933}.

\bibitem[{Dao and Gu(2024)}]{mamba2}
Dao, T.; and Gu, A. 2024.
\newblock Transformers are SSMs: generalized models and efficient algorithms through structured state space duality.
\newblock In \emph{Proceedings of the 41st International Conference on Machine Learning}, 10041--10071.

\bibitem[{Desai et~al.(2021)Desai, Freeman, Wang, and Beaver}]{timeVAE}
Desai, A.; Freeman, C.; Wang, Z.; and Beaver, I. 2021.
\newblock Timevae: A variational auto-encoder for multivariate time series generation.
\newblock \emph{arXiv preprint arXiv:2111.08095}.

\bibitem[{Godahewa et~al.(2021)Godahewa, Bergmeir, Webb, Hyndman, and Montero-Manso}]{monash}
Godahewa, R.; Bergmeir, C.; Webb, G.~I.; Hyndman, R.~J.; and Montero-Manso, P. 2021.
\newblock Monash time series forecasting archive.
\newblock \emph{arXiv preprint arXiv:2105.06643}.

\bibitem[{Goodfellow et~al.(2014)Goodfellow, Pouget-Abadie, Mirza, Xu, Warde-Farley, Ozair, Courville, and Bengio}]{gan}
Goodfellow, I.~J.; Pouget-Abadie, J.; Mirza, M.; Xu, B.; Warde-Farley, D.; Ozair, S.; Courville, A.; and Bengio, Y. 2014.
\newblock Generative adversarial nets.
\newblock \emph{Advances in neural information processing systems}, 27.

\bibitem[{Gu and Dao(2023)}]{mamba}
Gu, A.; and Dao, T. 2023.
\newblock Mamba: Linear-time sequence modeling with selective state spaces.
\newblock \emph{arXiv preprint arXiv:2312.00752}.

\bibitem[{Gu et~al.(2020)Gu, Dao, Ermon, Rudra, and R{\'e}}]{hippo}
Gu, A.; Dao, T.; Ermon, S.; Rudra, A.; and R{\'e}, C. 2020.
\newblock Hippo: Recurrent memory with optimal polynomial projections.
\newblock \emph{Advances in neural information processing systems}, 33: 1474--1487.

\bibitem[{Gu, Goel, and R{\'e}(2021)}]{s4}
Gu, A.; Goel, K.; and R{\'e}, C. 2021.
\newblock Efficiently modeling long sequences with structured state spaces.
\newblock \emph{arXiv preprint arXiv:2111.00396}.

\bibitem[{Gu et~al.(2021)Gu, Johnson, Goel, Saab, Dao, Rudra, and R{\'e}}]{lssl}
Gu, A.; Johnson, I.; Goel, K.; Saab, K.; Dao, T.; Rudra, A.; and R{\'e}, C. 2021.
\newblock Combining recurrent, convolutional, and continuous-time models with linear state space layers.
\newblock \emph{Advances in neural information processing systems}, 34: 572--585.

\bibitem[{Ho, Jain, and Abbeel(2020)}]{ddpm}
Ho, J.; Jain, A.; and Abbeel, P. 2020.
\newblock Denoising diffusion probabilistic models.
\newblock \emph{Advances in neural information processing systems}, 33: 6840--6851.

\bibitem[{Hu et~al.(2024)Hu, Baumann, Gui, Grebenkova, Ma, Fischer, and Ommer}]{zigma}
Hu, V.~T.; Baumann, S.~A.; Gui, M.; Grebenkova, O.; Ma, P.; Fischer, J.; and Ommer, B. 2024.
\newblock Zigma: A dit-style zigzag mamba diffusion model.
\newblock In \emph{European Conference on Computer Vision}, 148--166. Springer.

\bibitem[{Huang et~al.(2023)Huang, Shen, Zhang, Ding, Wang, Zhou, and Wang}]{crossgnn}
Huang, Q.; Shen, L.; Zhang, R.; Ding, S.; Wang, B.; Zhou, Z.; and Wang, Y. 2023.
\newblock Crossgnn: Confronting noisy multivariate time series via cross interaction refinement.
\newblock \emph{Advances in Neural Information Processing Systems}, 36: 46885--46902.

\bibitem[{Jeha et~al.(2022)Jeha, Bohlke-Schneider, Mercado, Kapoor, Nirwan, Flunkert, Gasthaus, and Januschowski}]{psa-gan}
Jeha, P.; Bohlke-Schneider, M.; Mercado, P.; Kapoor, S.; Nirwan, R.~S.; Flunkert, V.; Gasthaus, J.; and Januschowski, T. 2022.
\newblock PSA-GAN: Progressive self attention GANs for synthetic time series.
\newblock In \emph{The Tenth International Conference on Learning Representations}.

\bibitem[{Jeong et~al.(2025)Jeong, Sohn, Jeon, Shon, and Suk}]{freq-diff}
Jeong, S.; Sohn, J.; Jeon, J.; Shon, Y.; and Suk, H.-I. 2025.
\newblock Frequency-Conditioned Diffusion Models for Time Series Generation.

\bibitem[{Kingma and Welling(2022)}]{vae_bayes}
Kingma, D.~P.; and Welling, M. 2022.
\newblock Auto-Encoding Variational Bayes.
\newblock \emph{stat}, 1050: 10.

\bibitem[{Li et~al.(2025)Li, Meng, Bi, Urnes, and Chen}]{pad-ts}
Li, Y.; Meng, H.; Bi, Z.; Urnes, I.~T.; and Chen, H. 2025.
\newblock Population Aware Diffusion for Time Series Generation.
\newblock \emph{arXiv preprint arXiv:2501.00910}.

\bibitem[{Mogren(2016)}]{c_rnn_gan}
Mogren, O. 2016.
\newblock C-RNN-GAN: Continuous recurrent neural networks with adversarial training.
\newblock \emph{arXiv preprint arXiv:1611.09904}.

\bibitem[{Peebles and Xie(2023)}]{DiT}
Peebles, W.; and Xie, S. 2023.
\newblock Scalable diffusion models with transformers.
\newblock In \emph{Proceedings of the IEEE/CVF international conference on computer vision}, 4195--4205.

\bibitem[{Rangapuram et~al.(2018)Rangapuram, Seeger, Gasthaus, Stella, Wang, and Januschowski}]{ssm_intro}
Rangapuram, S.~S.; Seeger, M.~W.; Gasthaus, J.; Stella, L.; Wang, Y.; and Januschowski, T. 2018.
\newblock Deep state space models for time series forecasting.
\newblock \emph{Advances in neural information processing systems}, 31.

\bibitem[{Van~der Maaten and Hinton(2008)}]{tsne}
Van~der Maaten, L.; and Hinton, G. 2008.
\newblock Visualizing data using t-SNE.
\newblock \emph{Journal of machine learning research}, 9(11).

\bibitem[{W{\k{e}}glarczyk(2018)}]{kde}
W{\k{e}}glarczyk, S. 2018.
\newblock Kernel density estimation and its application.
\newblock In \emph{ITM web of conferences}, volume~23, 00037. EDP Sciences.

\bibitem[{Xiao et~al.(2024)Xiao, Li, Zhang, Meng, and Zhang}]{spatial-mamba}
Xiao, C.; Li, M.; Zhang, Z.; Meng, D.; and Zhang, L. 2024.
\newblock Spatial-Mamba: Effective Visual State Space Models via Structure-Aware State Fusion.
\newblock \emph{arXiv preprint arXiv:2410.15091}.

\bibitem[{Yoon, Jarrett, and Van~der Schaar(2019)}]{timeGAN}
Yoon, J.; Jarrett, D.; and Van~der Schaar, M. 2019.
\newblock Time-series generative adversarial networks.
\newblock \emph{Advances in neural information processing systems}, 32.

\bibitem[{Yuan and Qiao(2024)}]{diffusion-ts}
Yuan, X.; and Qiao, Y. 2024.
\newblock Diffusion-ts: Interpretable diffusion for general time series generation.
\newblock \emph{arXiv preprint arXiv:2403.01742}.

\bibitem[{Zeng et~al.(2023)Zeng, Chen, Zhang, and Xu}]{dlinear}
Zeng, A.; Chen, M.; Zhang, L.; and Xu, Q. 2023.
\newblock Are transformers effective for time series forecasting?
\newblock In \emph{Proceedings of the AAAI conference on artificial intelligence}, volume~37, 11121--11128.

\bibitem[{Zhou et~al.(2021)Zhou, Zhang, Peng, Zhang, Li, Xiong, and Zhang}]{informer}
Zhou, H.; Zhang, S.; Peng, J.; Zhang, S.; Li, J.; Xiong, H.; and Zhang, W. 2021.
\newblock Informer: Beyond efficient transformer for long sequence time-series forecasting.
\newblock In \emph{Proceedings of the AAAI conference on artificial intelligence}, volume~35, 11106--11115.

\bibitem[{Zhu et~al.(2023)Zhu, Ye, Zhang, Zhao, and Yu}]{difftraj}
Zhu, Y.; Ye, Y.; Zhang, S.; Zhao, X.; and Yu, J. 2023.
\newblock Difftraj: Generating gps trajectory with diffusion probabilistic model.
\newblock \emph{Advances in Neural Information Processing Systems}, 36: 65168--65188.

\end{thebibliography}

\clearpage

\appendix

\section{Additional Methodological Details}

\subsection{Notations}

For clarity and ease of understanding, the main notations in our method are summarized and presented in Table \ref{notation}.

\begin{table}[h]
\centering
\resizebox{1.0\linewidth}{!}{
\begin{tabular}{p{1.86cm}|p{5.9cm}}
\toprule
\textbf{Notations}                          & \multicolumn{1}{c}{\textbf{Descriptions}}                                                                            \\ \midrule
$L, C$                                      & temporal dimension and channel dimension of time series data                                                                        \\
$x^0$                                       & the sample from original data distribution                                                                           \\
$x^{T}$                                     & the sample from Gaussian noise                                                                                       \\
$x^t$                                       & the noised data at diffusion step $t$                                                                                \\
$\beta^{t}$                                 & a variance at diffusion step $t$                                                                                     \\
$t^{\text{emb}}$                                   & time embedding of diffusion step $t$                                                                                 \\
$x_k, y_k$                                  & the $k^{\text{th}}$ time step of input and output data                                                               \\
$h_k$                                       & the $k^{\text{th}}$ time step of latent state                                                                        \\
$u_k$                                       & the $k^{\text{th}}$ time step of lag fused latent state                                                              \\
$\Omega$                                    & lag set for latent state fusion                                                                                      \\
$l_p(k)$                                    & index of the $p^{\text{th}}$ lag of the $k^{\text{th}}$ time step                                                    \\
$\pi_i$                       & index of the $i^{\text{th}}$ channel in reordered permutation of CPS                   \\
$H$                                           & transformation matrix of permutation                                                                           \\
$\bar{\mathbf{A}}_i$                        & the $i^{\text{th}}$ state transition matrix                                                                          \\
$\bar{\mathbf{A}}_{i:k}^{\times}$ & product of the state transition matrices indexed from $i + 1$ to $k$  \\
$\bar{\mathbf{B}}_{i},\bar{\mathbf{C}}_{i}$ & the $i^{\text{th}}$ projection matrix                                                                                \\
$\mathbf{M},\mathbf{M}^F,\mathbf{M}^{C}$    & matrices in Mamba, Lag Fusion Mamba and Permutation Scanning Mamba under the matrix multiplication framework \\ \bottomrule
\end{tabular}}
\caption{Main notations used in article.}
\label{notation}
\end{table}

\subsection{Motivation}

\begin{figure}[h]
\centering
\includegraphics[width=0.95\linewidth]{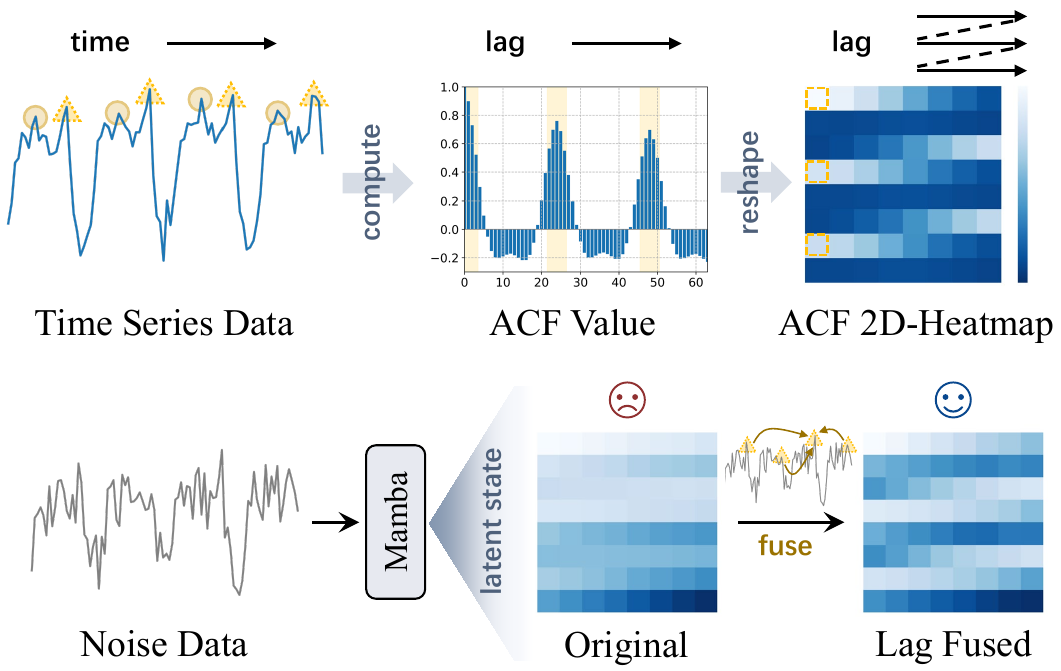}
\caption{Illustration of a motivating example.}
\label{app_intro}
\end{figure}

Periodicity is a fundamental characteristic of time series data. As shown in Figure \ref{app_intro}, the autocorrelation coefficient reveals the similarity between the current time step and periodical lag. However, due to the inherent constraint of unidirectional scanning paradigm of SSMs, it is challenging to retain such temporal dependencies over long-range periodic intervals. Our Lag Fusion Mamba explicitly fuses lag state in latent state space to mitigates the issue.

\section{Denoising Diffusion Probabilistic Models}
\label{DDPM}

In tis section, we provide a overview of DDPMs. The framework comprises two phases: a forward diffusion process that incrementally adds noise to the data according to a predefined schedule, and a reverse denoising process modeled by deep neural networks to reconstruct the original input.

In forward process, given a data distribution $q(x)$ and sample initial $x^0 \sim q(x^0)$ from it. In the diffusion process up to step $T$, a sequence of progressively noised latent variables $x^1, x^2,\ldots ,x^T$ is generated via a Markov chain:
\begin{equation}
    q(\mathbf{x}^t|\mathbf{x}^{t-1}):=\mathcal{N}(\mathbf{x}^t;\sqrt{1-\beta^t}\mathbf{x}^{t-1},\beta^t\mathbf{I}),
\end{equation}
where $\beta^t \in (0,1)$ is a variance at diffusion step $t$. According to the properties of the Gaussian kernel, it is feasible to get noise samples directly from original input $x^0$:
\begin{equation}
    q(\mathbf{x}^t|\mathbf{x}^0):=\prod_{t=1}^Tq(\mathbf{x}^t|\mathbf{x}^{t-1}):=\mathcal{N}(\mathbf{x}^t;\sqrt{\bar{\alpha}^t}\mathbf{x}^0,\sqrt{1-\bar{\alpha}^t}\mathbf{I}),
\end{equation}
where $\alpha^t:=1-\beta^t$ and $\bar{\alpha}:=\prod_{i=1}^T\alpha^i$. Using reparameterization trick and defining $\epsilon\sim\mathcal{N}(0,\mathbf{I})$, we have:
\begin{equation}
    x^t=\sqrt{\bar{\alpha}^t}x^0+\sqrt{1-\bar{\alpha}^t}\epsilon.
\end{equation}

In the reverse process, the series of reverse Markov chains start from a distribution $p(x^T)=\mathcal{N}(x^T;\mathbf{0},\mathbf{I})$ and the noise $x^T$. We can utilize model $p_{\theta}(x^{t-1}|x^{t})$ to get $x^0$. The diffusion model is trained to minimize variational constraints on the negative log-likelihood (NLL), which is equivalent to Kullback–Leibler (KL) divergence format:
\begin{equation}
    \begin{aligned}
L:= & \mathbb{E}_{q}[\underbrace{-\log p_\theta\left({x}^0\mid x^1\right)}_{L_0}+\underbrace{D_{\mathrm{KL}}\left(q\left({x}^{T}\mid{x}^{0}\right)\parallel p\left({x}^{T}\right)\right)}_{L_{T}} \\
 & +\sum_{t>1}\underbrace{D_{\mathrm{KL}}\left(q\left({x}^{t-1}\mid {x}^{t},{x}^{0}\right)\parallel p_{\theta}\left({x}^{t-1}\mid {x}^{t}\right)\right)}_{L_{t-1}}].
\end{aligned}
\end{equation}
Subsequently, to minimize the divergence loss $\mathcal{L}_{t-1}$, Bayes’ rule is applied to parameterize posterior $q\left({x}^{t-1}\mid{x}^{t},{x}^{0}\right)$:
\begin{equation}
\begin{array}{cc}
    q\left({x}^{t-1}\mid{x}^{t},{x}^{0}\right)=\mathcal{N}\left({x}^{t-1};\tilde{\mu}^{t}\left({x}^{t},{x}^{0}\right),\frac{1-\bar{\alpha}^{t-1}}{1-\bar{\alpha}^t}\beta^t\mathbf{I}\right),\\
    \tilde{\mu}^{t}({x}^{t},{x}^{0}):=\frac{\sqrt{\bar{\alpha}^{t-1}}\beta^{t}}{1-\bar{\alpha}^{t}}{x}^{0}+\frac{\sqrt{\alpha^{t}}(1-\bar{\alpha}^{t-1})}{1-\bar{\alpha}^{t}}{x}^{t}.
\end{array}
\end{equation}
Finally, it can be equated to predicting noise $\epsilon$:
\begin{equation}
    \mathcal{L}=\mathbb{E}_{t,x^0,\epsilon}\left[\lambda(t)\left\|\epsilon-\epsilon_\theta(\mathbf{x}^t,t)\right\|^2\right],
\end{equation}
where $\lambda(t) =\frac{(\beta^t)^2}{2\sigma_T^2\alpha^t(1-\bar{\alpha}^t)}$ is a noise weight.

In sampling stage, given the noisy $x^t$ and step $t$, the reverse process $p_\theta(x^{t-1}|x^t)$ can be written as:
\begin{equation}
    x^{t-1}=\frac{1}{\sqrt{\alpha^t}}(x^t-\frac{\beta^t}{\sqrt{1-\bar{\alpha}^t}}\epsilon_\theta(x^t,t))+\sigma^tz,
\end{equation}
where $z\sim\mathcal{N}(0,\mathbf{I})$ is a standard Gaussian noise, $\sigma^t$ is hyperparameter set to $\beta^t$ in practice.

\section{DiFM and DiPM}\label{DiFM_DiPM}

\begin{figure}[t]
\centering
\includegraphics[width=0.85\linewidth]{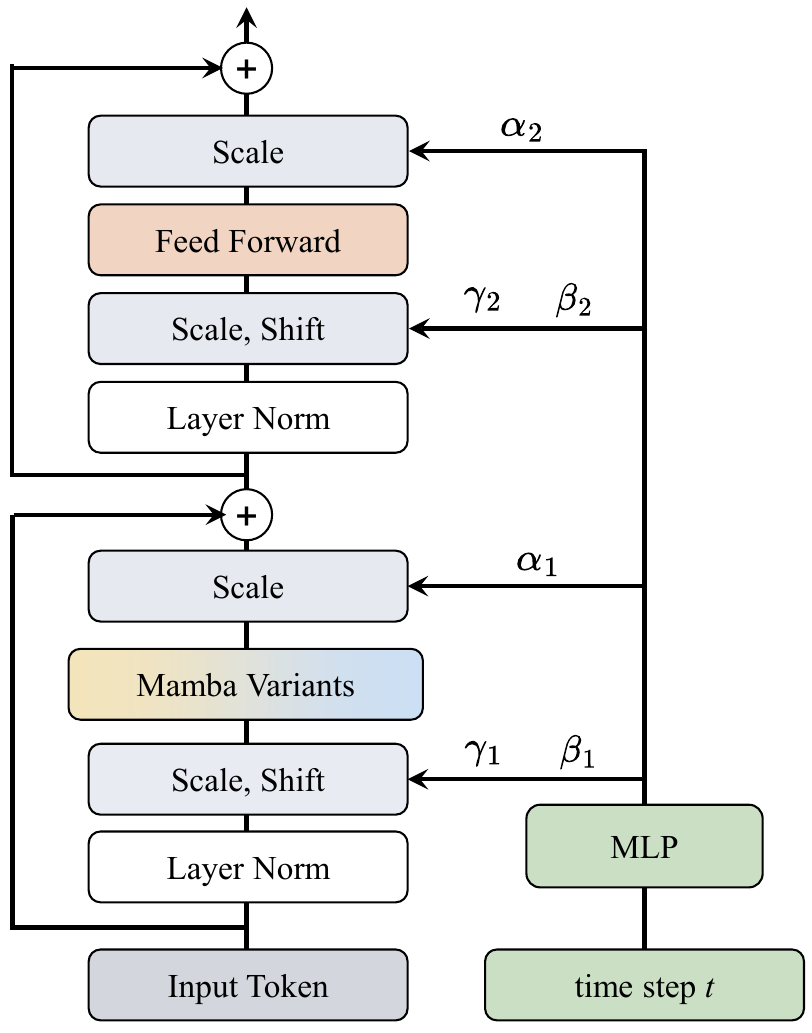}
\caption{Architectures of DiFM and DiPM.}
\label{DiM}
\end{figure}

In this section, we present the structure of Diffusion Fusion Mamba (DiFM) and Diffusion Permutation Mamba (DiPM) in DiM-TS. As shown in Figure \ref{DiM}, both architectures mirror the backbone of DiT \cite{DiT}, with the multi-head attention block replaced by our core Mamba variants — Lag Fusion Mamba and Permutation Scanning Mamba. We exclusively employ the diffusion timestep $t$ as the condition input, embedding it through a MLP network to guide the generation process. Specifically, condition embedding feature $t^{emb}$ is divided into six chunks: $\alpha_1,\beta_1,\gamma_1,\alpha_2,\beta_2,\gamma_2$, each chunk is then incorporated into scale and shift layers. Such architecture design has demonstrated strong scalability and promising generation quality.

\section{Derivation of SSMs Formulas}\label{SSMs}

\subsection{Mamba}

Based on the Mamba formulation defined in Eq. (\ref{eq_o}), the state transition equation can be reformulated as recursive form:
\begin{equation}\label{Mamba_derivation}
\begin{aligned}
h_{k} & =\bar{\mathbf{A}}_{k}h_{k-1}+\bar{\mathbf{B}}_{k}x_{k} \\
 & =\bar{\mathbf{A}}_k\left(\bar{\mathbf{A}}_{k-1}h_{k-2}+\bar{\mathbf{B}}_{k-1}x_{k-1}\right)+\bar{\mathbf{B}}_kx_k \\
 & =\bar{\mathbf{A}}_k\bar{\mathbf{A}}_{k-1}h_{k-2}+\bar{\mathbf{A}}_k\bar{\mathbf{B}}_{k-1}x_{k-1}+\bar{\mathbf{B}}_kx_k \\
 & =\Pi_{i=1}^k\bar{\mathbf{A}}_ih_0+\Pi_{i=2}^k\bar{\mathbf{A}}_i\bar{\mathbf{B}}_1x_1+\cdots\\
 & \quad +\Pi_{i=t}^k\bar{\mathbf{A}}_i\bar{\mathbf{B}}_{t-1}x_{t-1}+\bar{\mathbf{B}}_tx_t
\end{aligned}.
\end{equation}
By setting state $h_0=\bar{\mathbf{{B}}}_0x_0$,  Eq. (\ref{Mamba_derivation}) can be rewritten as:
\begin{equation}
h_k=\sum_{i=0}^{k}\bar{\mathbf{A}}_{i:k}^{\times}\bar{\mathbf{B}}_{i}x_i, \quad\bar{\mathbf{A}}_{i:k}^\times:=
\begin{cases}
\Pi_{j=i+1}^k\bar{\mathbf{A}}_j, & i<k \\
1, & i=k
\end{cases}.
\end{equation}
According to the observation equation, the output $y_k$ is obtained by multiplying $\mathbf{C}_k$ to the latent state:
\begin{equation}
y_k=\sum_{i=0}^{k}\mathbf{C}_k^{\top}\bar{\mathbf{A}}_{i:k}^{\times}\bar{\mathbf{B}}_{i}x_i .
\end{equation}
Vectorizing the input over $k$, i.e., $x=[x_0,\ldots,x_k]$, the corresponding output vector is $y=[y_0,\ldots,y_k]$. Then, we establish the matrix transformation form of SSMs:$
y=\mathbf{M}x$,
where $\mathbf{M}$ is a lower triangular matrix, $\mathbf{M}_{ki}=\mathbf{C}_k^{\top}\bar{\mathbf{A}}_{i:k}^{\times}\bar{\mathbf{B}}_{i}$.

\subsection{Lag Fusion Mamba}

By following Mamba derivation, LSF can be expressed as:
\begin{equation}
u_k=\sum_{p\in\Omega}\eta_ph_{l_p(k)}=\sum_{p\in\Omega}\sum_{i\leq l_p(k)}\eta_p\bar{\mathbf{A}}_{i:l_p(k)}^{\times}\bar{\mathbf{B}}_{i}x_i,
\end{equation}
Multiplying by $\mathbf{C}_k$, the output of LSF can be derived:
\begin{equation}
y_k=\sum_{p\in\Omega}\sum_{i\leq l_p(k)}\eta_p\mathbf{C}_k\bar{\mathbf{A}}_{i:l_p(k)}^{\times}\bar{\mathbf{B}}_{i}x_i.
\end{equation}
Similarly, it can be can be formalized as a matrix multiplication form $y=\mathbf{M}^{F}x$, where $\mathbf{M}^{F}$ is a adjacency matrix, $\mathbf{M}^F_{ki}=\sum_{p\in\Omega}\eta_p\mathbf{C}_k^{\top}\bar{\mathbf{A}}_{i:l_p(k)}^{\times}\bar{\mathbf{B}}_{i}$.

\subsection{Permutation Scanning Mamba}

We firstly derive the transformation matrix $H$. The objective function in Eq. (\ref{laplace}) can be expanded to its simplified form:
\begin{equation}
\begin{aligned}
& \frac{1}{2} \sum_{i=1}^{C} \sum_{j=1}^{C}\left\|z_{i}-z_{j}\right\|^{2} g_{i j} \\
= & \frac{1}{2} \sum_{i=1}^{C} \sum_{j=1}^{C}\left(z_{i}^{\top} z_{i} g_{i j}-2 z_{i} z_{j} g_{i j}+z_{j}^{\top} z_{j} g_{i j}\right) g_{i j} \\
= & \sum_{i=1}^{C} \mathbf{D}_{i i} z_{i}^{\top} z_{i}-\sum_{i=1}^{C} \sum_{j=1}^{C} z_{i} z_{j} g_{i j} \\
= & \operatorname{tr}\left(Z^{\top} \mathbf{L} Z\right)
\end{aligned},
\end{equation}
where $V=(v_1,v_2,\ldots,v_C)^{\top}$ is channel mapping vector, $\mathbf{L}$ is the Laplacian matrix of $\mathbf{G}$, $\text{tr}(\cdot)$ represents the matrix trace. To ensure that mapped value is distributed throughout space rather than converging, additional constraint $V^{\top}\mathbf{D}V=\mathbf{I}$ is introduced, where $\mathbf{D}$ is the degree matrix of $\mathbf{G}$. Then the Lagrange multipliers is utilized:
\begin{equation}
\begin{array}{l}
f(Z)=\operatorname{tr}\left(Z^\top \mathbf{L} Z\right)+\operatorname{tr}\left[\mathbf{\Lambda}\left(Z^\top \mathbf{D} Z-\mathbf{I}\right)\right], \frac{\partial f(Z)}{\partial Z}=0\\
\Rightarrow \mathbf{L}Z+\mathbf{L}^{\top} Z+\mathbf{D}^{\top}Z \mathbf{\Lambda}^{\top}+\mathbf{D} Z \mathbf{\Lambda}=0 \\
\Rightarrow \mathbf{L} Z=-\mathbf{D} Z \mathbf{\Lambda}
\end{array},
\end{equation}
where $\mathbf{\Lambda}$ is a diagonal matrix. The desired vector $V$ is obtained by computing the eigenvector corresponding to the smallest non-zero eigenvalue. The elements are then sorted in order to obtain the ordered vector $V_{\pi}=(v_{\pi_1},v_{\pi_2},\ldots,v_{\pi_C})$, where $v_{\pi_1}<v_{\pi_2}<\ldots<v_{\pi_C}$. This induces the target permutation $\pi=\{\pi_1,\pi_2,\ldots,\pi_C\}$. By defining $H_{ij}=\mathds{1}_{i=\pi_j}$, the transformation matrix $H$ is constructed to achieve the desired channel permutation.

To enable the model to scan the channels in a specific permutation as defined in Eq. (\ref{ssm_p}), the input is transformed into $Hx=[x_{\pi_1},\ldots,x_{\pi_C}]$, followed by standard Mamba:
\begin{equation}
    h_{\pi_k}=\sum_{i=0}^{k}\bar{\mathbf{A}}_{{\pi_i}:{\pi_k}}^{\times}\bar{\mathbf{B}}_{{\pi_i}}x_{\pi_i}, \text{ }y_{\pi_k}=\sum_{i=0}^{k}\mathbf{C}_{\pi_k}^{\top}\bar{\mathbf{A}}_{{\pi_i}:{\pi_k}}^{\times}\bar{\mathbf{B}}_{{\pi_i}}x_{\pi_i}.
\end{equation}
It can also be equivalently represented in the form of matrix multiplication: $y=\mathbf{M}(Hx)$. Finally, the output $y=[y_{\pi_1},\ldots,y_{\pi_C}]$ is mapped back to the original order by the inverse of transformation matrix, $H^{-1}$, which exists due to the bijective nature of permutation. The overall process can be formulated as $y=\mathbf{M}^Cx$, where $\mathbf{M}^C=H^{-1}\mathbf{M}H$.

\section{Experiment Details}\label{exp_detail}

\subsection{Datasets}

Table \ref{statistic} lists the statistics of the datasets in our experiments. These datasets are all available online \cite{monash,energy,informer}.

\begin{table}[h]
\centering
\caption{The statistics of benchmark datasets.}
\begin{tabular}{cccc}
\toprule
Dataset & Samples & Channels & Granularity \\ \midrule
Stocks  & 3685    & 6        & 1 day       \\
ETTh    & 17420   & 7        & 1 hour      \\
Energy  & 19735   & 28       & 10 min      \\
KDD-Cup & 10920   & 59       & 1 hour      \\
Traffic & 48204   & 1        & 1 hour      \\ \bottomrule
\end{tabular}
\label{statistic}
\end{table}

\begin{table*}[t]
\centering
\begin{tabular}{>{\centering\arraybackslash}m{2.8cm}|>{\centering\arraybackslash}m{2.18cm}>{\centering\arraybackslash}m{2.18cm}>{\centering\arraybackslash}m{2.18cm}>{\centering\arraybackslash}m{2.18cm}>{\centering\arraybackslash}m{2.18cm}}
\toprule
Metrics           & DiM-TS                     & PaD-TS            & Diffusion-TS               & TimeVAE                    & TimeGAN \\ \midrule
Context-FID score & \textbf{0.1846$\pm$0.0296} & 1.3713$\pm$0.1509 & 0.3096$\pm$0.0330          & 0.4250$\pm$0.0672          & -       \\ \midrule
Predictive score  & \textbf{0.2395$\pm$0.0001} & 0.2399$\pm$0.0001 & \textbf{0.2395$\pm$0.0001} & \textbf{0.2395$\pm$0.0001} & -       \\ \bottomrule
\end{tabular}
\caption{Generation results with length 64 on Traffic dataset. '-' indicates that the model failed to converge during training.}
\label{traffic}
\end{table*}

\subsection{Evalutation Metrics}\label{eva_metric}

\noindent\textbf{1) Context-FID score}
It replace the original FID metric with a time series representation learned by TS2Vec \cite{diffusion-ts}. A lower score indicates that the synthetic data distribution is closer to the original data, which is generally more beneficial for downstream forecasting performance. Specifically, synthetic time series and real-time series are first encoded into representations using a pre-trained TS2Vec model, followed by computation of the FID score.

\noindent\textbf{2) Correlational score}
We first estimate the covariance of $i^{\text{th}}$ and $j^{\text{th}}$ channel of time series:
\begin{equation}
    \operatorname{cov}_{i, j}=\frac{1}{L} \sum_{t=1}^{L} x_{i}^{t} x_{j}^{t}-\left(\frac{1}{L} \sum_{t=1}^{L} x_{i}^{t}\right)\left(\frac{1}{L} \sum_{t=1}^{L} x_{j}^{t}\right) .
\end{equation}
Then the difference of covariance between the synthetic data and original data is computed by:
\begin{equation}
    \frac{1}{10} \sum_{i, j}^{d}\left|\frac{\operatorname{cov}_{i, j}^{r}}{\sqrt{\operatorname{cov}_{i, i}^{r}} \operatorname{cov}_{j, j}^{r}}-\frac{\operatorname{cov}_{i, j}^{f}}{\sqrt{\operatorname{cov}_{i, i}^{f}} \operatorname{cov}_{j, j}^{f}}\right|.
\end{equation}

\noindent\textbf{3) Discriminative score}
It is based on a $2$-layer GRU-based classifier (clf) trained with the training set from original and synthetic datasets (with label synthetic$=0$, real$=1$). Then the metric is calculated on the test sets:
\begin{equation}
\left|\frac{\sum_{n=1}^S(0=\mathrm{clf}({x}_n^f))+\sum_{n=1}^S(1=\mathrm{clf}(x_n^r))}{2S}-0.5\right|,
\end{equation}
where $S$ is set length, $x_n^r$ is real data, ${x}_n^f$ is synthetic data.

\noindent\textbf{4) Predictive score}
It employs the same GRU-based neural network architecture as used in Discriminative score. The model is evaluated by computing the Mean Absolute Error (MAE) between the ground-truth and predicted values, using the generated data as input.

\noindent\textbf{5) Marginal Distribution Difference}
For each dimension and time step, this metric constructs histograms of the generated series using the bin centers and widths from the original series. The average absolute bin-wise difference is then computed to quantify the distributional alignment between the original and generated series.

\noindent\textbf{6) AutoCorrelation Difference}
Following \cite{tsgbench}, it computes and contrasts the autocorrelation of the original and generated time series to evaluate the preservation of temporal dependencies.

\noindent\textbf{7) Skewness Difference}
It is the statistical measure that quantify the distribution asymmetry of time series. Specifically, given the mean and standard deviation of the original time series ($\mu^{r}$, $\sigma^{r}$) and generated time series ($\mu^{f}$, $\sigma^{f}$), we evaluate the fidelity of generation by skewness difference:
\begin{equation}
\mathrm{SD}=\left|\frac{\mathbb{E}[(x^r-\mu^{r})^{3}]}{(\sigma^{r})^3}-\frac{\mathbb{E}[({x}^f-\mu^{f})^{3}]}{(\sigma^{f})^3}\right|.
\end{equation}

\noindent\textbf{8) Kurtosis Difference}
Similar to skewness, Kurtosis assesses the tail behavior of a distribution, capturing the presence of extreme deviations from the mean. The kurtosis difference is computed as:
\begin{equation}
\mathrm{KD}=\left|\frac{\mathbb{E}[(x^r-\mu^{r})^{4}]}{(\sigma^{r})^4}-\frac{\mathbb{E}[({x}^f-\mu^{f})^{4}]}{(\sigma^{f})^4}\right|.
\end{equation}

\noindent\textbf{9) Value Distribution Shift}
It measure the population-level distribution shift of generated time series in terms of value \cite{pad-ts} as follows:
\begin{equation}
    \mathrm{VDS}=\frac{1}{C}\sum_{i=1}^CD(P_V^i,Q_V^i),
\end{equation}
where $D$ denotes a certain distribution distance measure (e.g., KL divergence), $P^i_V$ is the value distribution of the $i^{\text{th}}$ channel in real data, and $Q^i_V$ is the corresponding distribution for the synthetic data.

\noindent\textbf{10) Functional Dependency Distribution Shift} It considers population-level distribution shift in terms of functional dependency by capturing the cross-correlation distribution across dataset as follows:
\begin{equation}
    \mathrm{FDDS}=\frac{1}{P}\sum_{p=1}^PD(P_{\mathrm{FD}}^{i,j},Q_{\mathrm{FD}}^{i,j}),
\end{equation}
where $P$ is the number of all pairs of channels, $P_{\mathrm{FD}}^{i,j}$ is the distribution of the functional dependency scores between $i^{\text{th}}$ and $j^{\text{th}}$ channel calculated by cross-correlation over the original data. $Q_{\mathrm{FD}}^{i,j}$ is the distribution of synthetic data.

\subsection{Hyperparameter Setting}

The hyperparameters of DiM-TS can be divided into two categories: the diffusion model pipeline and the model architecture. They are summarized in Table \ref{Hyperparameters}. Detailed configurations can be found in code. The hyperparameter sensitivity analysis are presented in the subsequent section \ref{Hyper_Analysis}.

\begin{table}[h]
\centering
\caption{List of model-related parameters.}
\resizebox{1.0\linewidth}{!}{
\begin{tabular}{>{\centering\arraybackslash}m{2.4cm}|>{\centering\arraybackslash}m{1.5cm}>{\centering\arraybackslash}m{1.5cm}>{\centering\arraybackslash}m{1.5cm}>{\centering\arraybackslash}m{1.5cm}}
\toprule
Parameter        & Stocks               & ETTh                 & Energy               & KDD-Cup              \\ \midrule
Target           & $x^0$ & $x^0$ & $x^0$ & $x^0$ \\
Noise Scheduler  & Cosine               & Cosine               & Cosine               & Cosine               \\
Batch Size       & 64                   & 128                  & 64                   & 128                  \\
Diffusion Step   & 500                  & 500                  & 500                  & 500                  \\
Normalization    & {[}-1,1{]}           & {[}-1,1{]}           & {[}-1,1{]}           & {[}-1,1{]}           \\ \midrule
Hidden Dim       & 128                  & 128                  & 256                  & 256                  \\
Dilation Factors & {[}1,2,3{]}          & {[}1,2,3{]}          & {[}1,2{]}            & {[}1,2,3{]}          \\
Num of Enc       & 1                    & 1                    & 1                    & 1                    \\
Num of DiFM      & 3                    & 3                    & 3                    & 3                    \\
Num of DiPM      & 3                    & 3                    & 3                    & 3                    \\
\bottomrule
\end{tabular}}
\label{Hyperparameters}
\end{table}

\section{Additional Experimental Results}\label{add_exp}

In this section, we present additional experimental results omitted in the main body of the paper due to limited space.

\subsection{Evaluation on Traffic Dataset}

Since the dataset comprises univariate time series with pronounced periodic patterns, certain evaluatio nmetrics are not applicable in this context. Therefore, we omit them and report the results in Table \ref{traffic}.

\begin{figure}[h]
  \centering
  \includegraphics[width=0.98\linewidth]{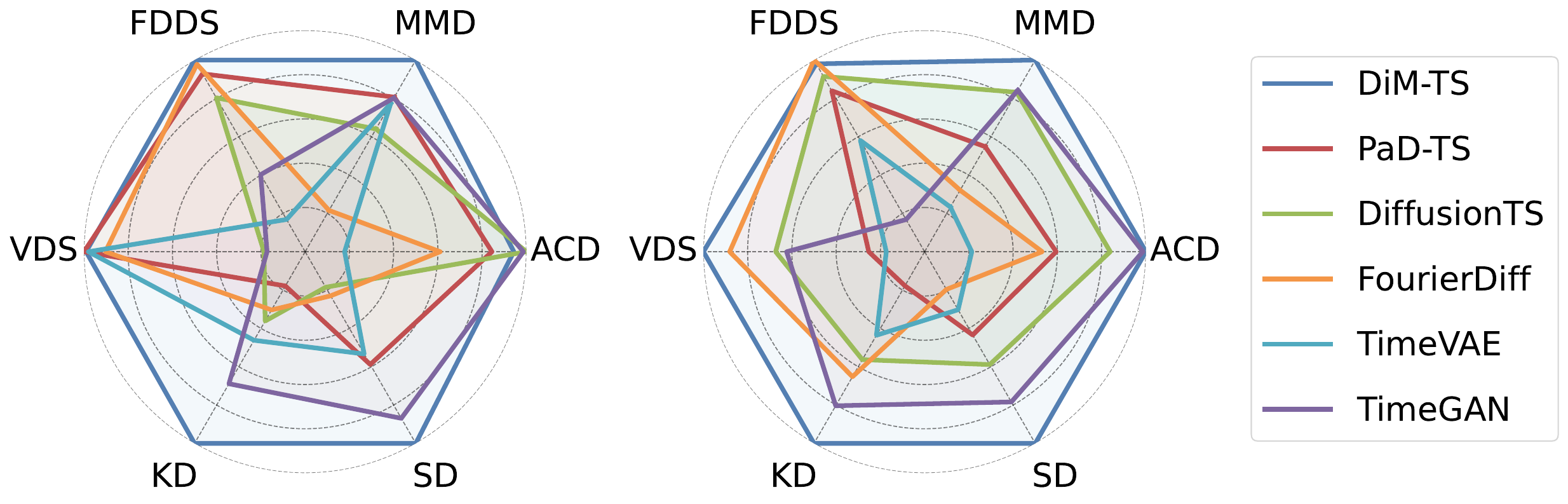}
  \caption{Feature-based and population-level measures comparison on Stocks (left) and ETTh (right).}
  \label{radar2}
\end{figure}

\subsection{Evaluation on Additional Metrics}

Figure \ref{radar2} compares the DiM-TS performance with baselines on the Stocks and ETTh datasets across additional metrics. It shows that DiM-TS achieves optimal performance across various settings.

\subsection{Hyperparameter Sensitivity Analysis}
\label{Hyper_Analysis}

\begin{figure}[t]
        \centering
        \begin{subfigure}{\linewidth}
		\centering
		\includegraphics[width=\linewidth]{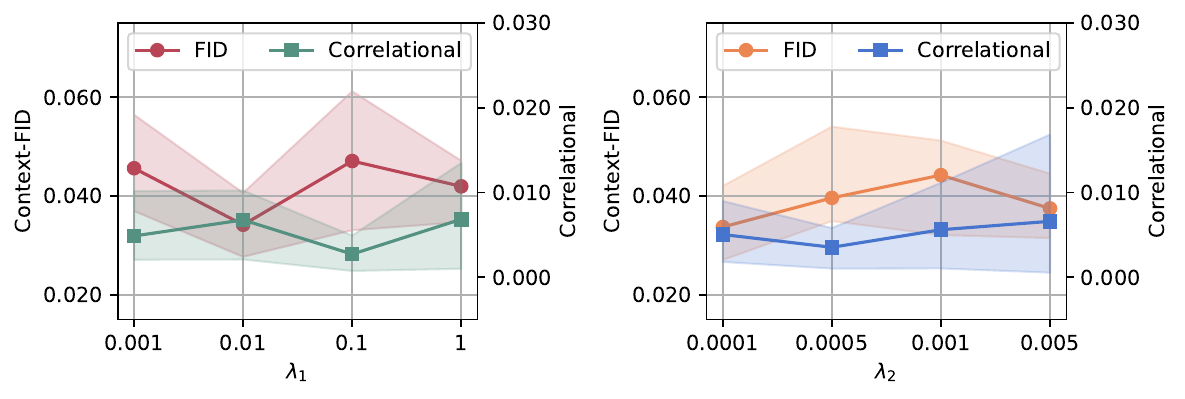}
		\caption{Stocks}
		\label{hyper_stock}
	\end{subfigure}
	\begin{subfigure}{\linewidth}
		\centering
		\includegraphics[width=\linewidth]{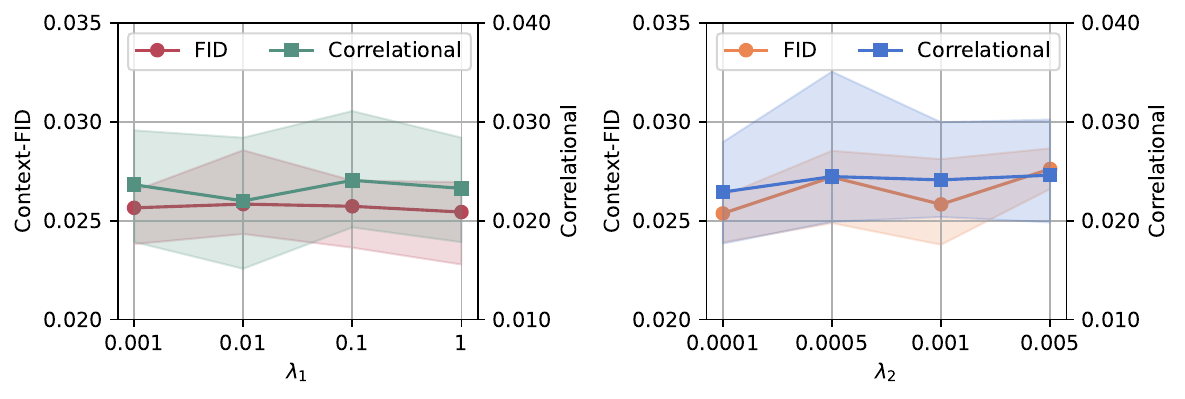}
		\caption{ETTh}
		\label{hyper_etth}
	\end{subfigure}
        \begin{subfigure}{\linewidth}
		\centering
		\includegraphics[width=\linewidth]{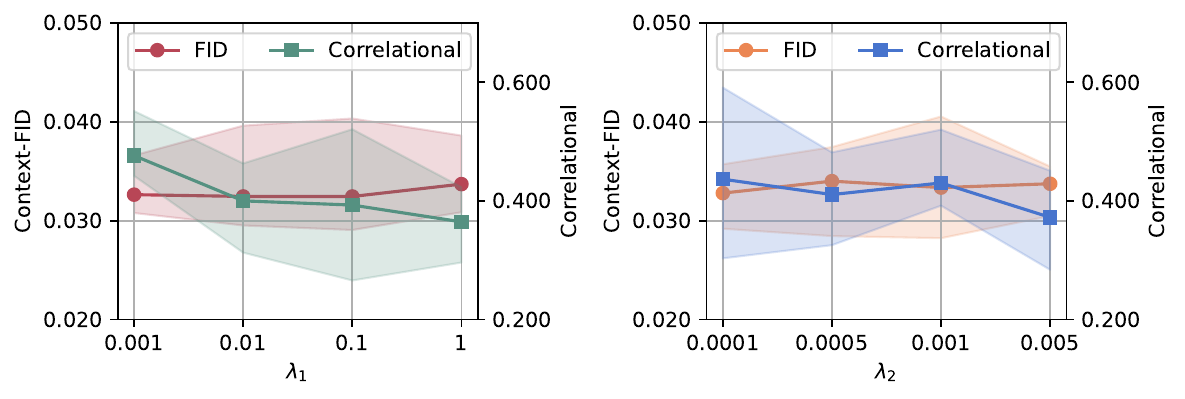}
		\caption{Energy}
		\label{hyper_energy}
	\end{subfigure}
\caption{Analysis of $\lambda_1$ and $\lambda_1$.}
\label{hyper_lambda}
\end{figure}

We investigated the impact of hyperparameter $\lambda_1$ and $\lambda_2$ that control the Fourier-based auxiliary loss and correlation regularization loss, respectively. The results are reported in Figure \ref{hyper_lambda}. It can be observed that different datasets and evaluation metrics may require tailored hyperparameter settings to achieve optimal performance. This provides guidance for hyperparameter selection in experiments.

\subsection{Additional Visualizations}\label{add_vis}

\begin{figure*}[htbp]
	\centering
	\begin{subfigure}{\linewidth}
		\centering
		\includegraphics[width=\linewidth]{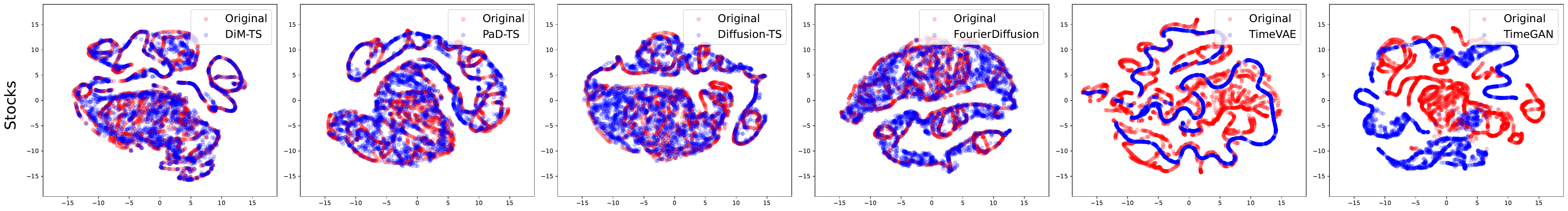}
	\end{subfigure}
        \centering
	\begin{subfigure}{\linewidth}
		\centering
		\includegraphics[width=\linewidth]{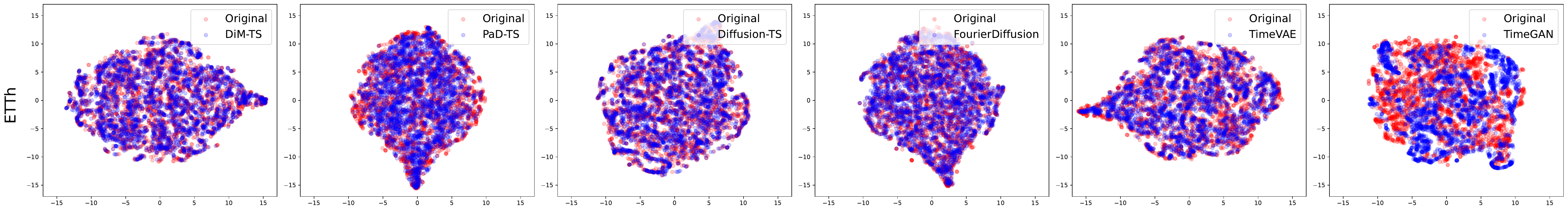}
	\end{subfigure}
        \centering
	\begin{subfigure}{\linewidth}
		\centering
		\includegraphics[width=\linewidth]{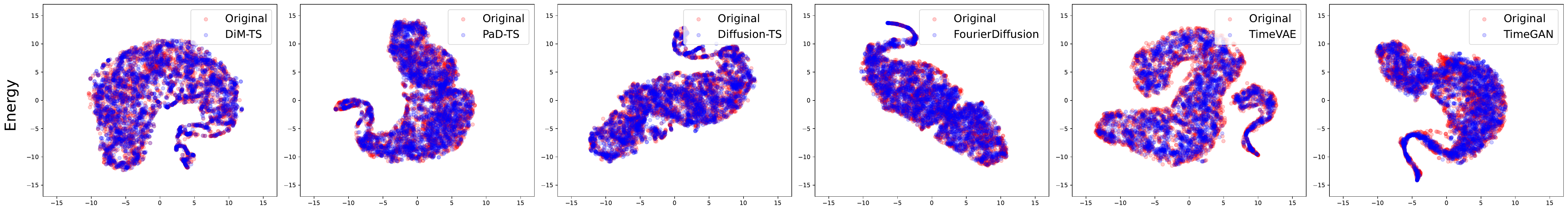}
	\end{subfigure}
        \centering
	\begin{subfigure}{\linewidth}
		\centering
		\includegraphics[width=\linewidth]{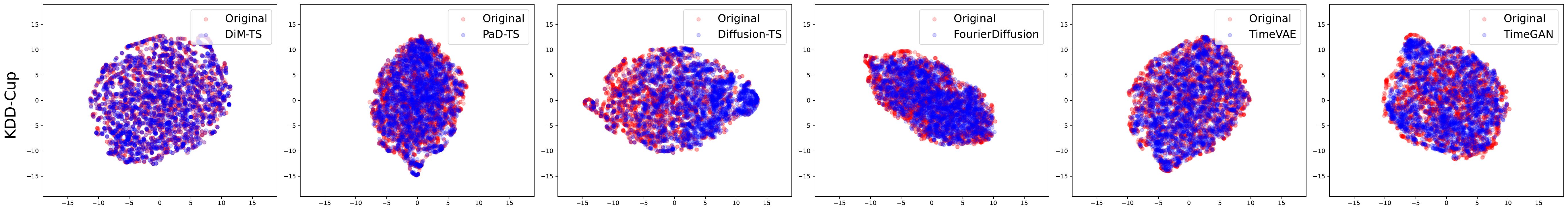}
	\end{subfigure}
\caption{t-SNE visualizations of all methods between original data (red) and synthetic data (blue) across datasets.}
\label{tsne}
\end{figure*}

\begin{figure*}[htbp]
	\centering
	\begin{subfigure}{\linewidth}
		\centering
		\includegraphics[width=\linewidth]{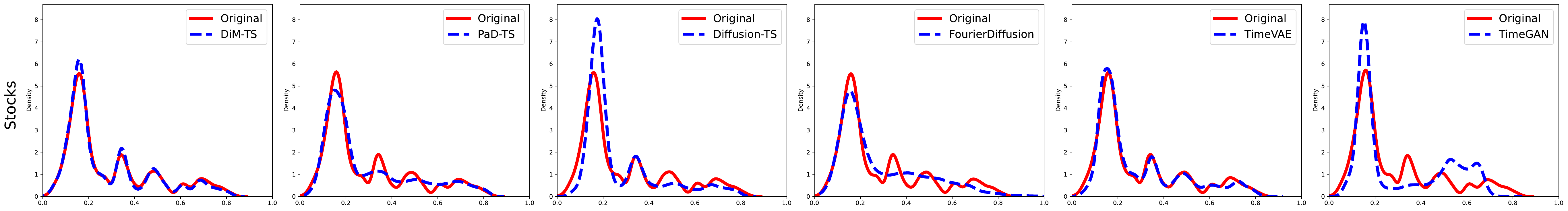}
	\end{subfigure}
        \centering
	\begin{subfigure}{\linewidth}
		\centering
		\includegraphics[width=\linewidth]{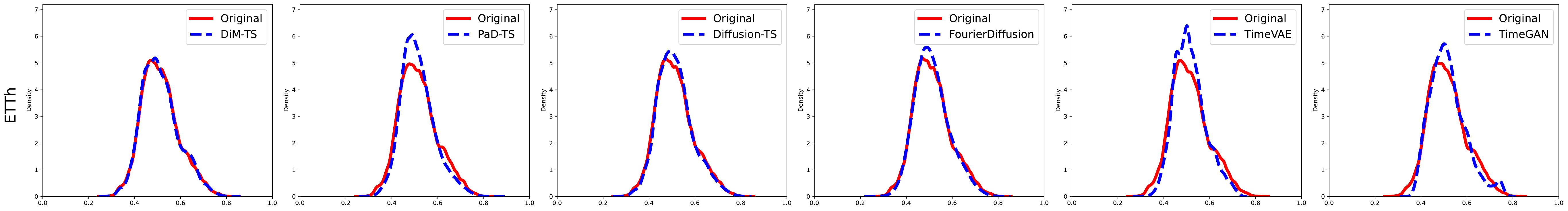}
	\end{subfigure}
        \centering
	\begin{subfigure}{\linewidth}
		\centering
		\includegraphics[width=\linewidth]{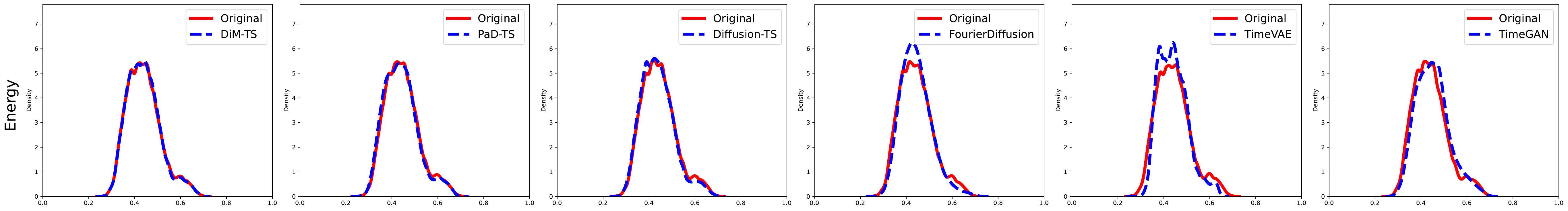}
	\end{subfigure}
        \centering
	\begin{subfigure}{\linewidth}
		\centering
		\includegraphics[width=\linewidth]{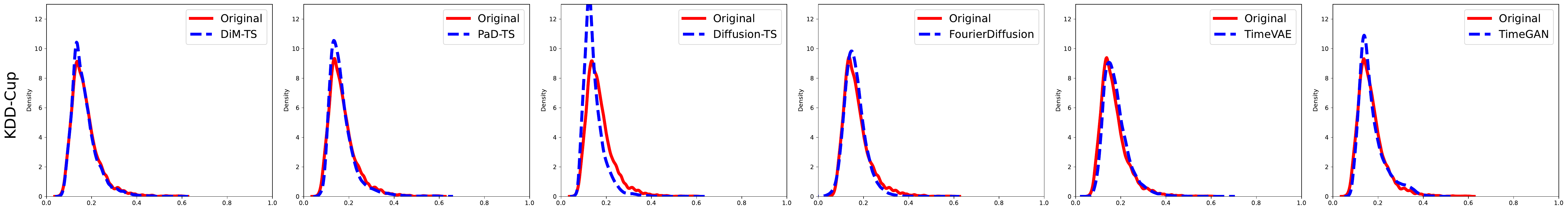}
	\end{subfigure}
\caption{Value distribution of all methods between original data (red) and synthetic data (blue) across datasets.}
\label{kernel}
\end{figure*}

We present additional t-SNE visualization and distribution outcomes in Figure \ref{tsne} and Figure \ref{kernel}.

\end{document}